\documentclass[10pt,journal,compsoc]{IEEEtran}
\ifCLASSOPTIONcompsoc
\usepackage[nocompress]{cite}
\else
\usepackage{cite}
\fi
\usepackage{amsmath}
\usepackage{amssymb}
\usepackage{bbm}
\usepackage{enumitem}
\usepackage{epsfig}
\usepackage{graphicx}
\usepackage{amsmath}
\usepackage{amssymb}
\usepackage{enumitem}
\usepackage{xcolor}
\usepackage{hyperref}
\usepackage{times}
\usepackage{boldline} 
\usepackage{multirow}
\usepackage{array, booktabs}
\usepackage[export]{adjustbox}

\graphicspath{{../img/}}
\newcommand{\Tau}{\mathcal{T}}

\newcommand{\thickhline}{%
	\noalign {\ifnum 0=`}\fi \hrule height 1pt
	\futurelet \reserved@a \@xhline
}

\newif\ifshowdiff
\showdifffalse 

\newcommand{\new}[1]{
	\ifshowdiff
	\textcolor{black}{#1}
	\else
	{#1}
	\fi }

\newcommand{\edited}[1]{
	\ifshowdiff
	\textcolor{teal}{#1}
	\else
	{#1}
	\fi }

\newcommand{\old}[1]{
	\ifshowdiff
	\textcolor{red}{#1}
	\fi }

\newif\ifrefinement
\refinementfalse 

\newcommand{\etal}{\textit{et al.}}
\graphicspath{{./}}

\setitemize[0]{leftmargin=10pt}

\newcommand{\revision}[1]{{\color{black}{#1}}}
\newcommand{\minorrevision}[1]{{\color{black}{#1}}}

\ifCLASSINFOpdf
\else
\fi
\begin{document}
%
\title{SMEMO: Social Memory for Trajectory Forecasting}
%
%
%
%

\author{Francesco Marchetti,
		Federico Becattini,
		Lorenzo Seidenari,
		Alberto Del Bimbo
\IEEEcompsocitemizethanks{\IEEEcompsocthanksitem F. Marchetti, F. Becattini, L. Seidenari, A. Del Bimbo are with the Media Integration and Communication Center (MICC) of the University of Florence, Italy.
}
}

\IEEEtitleabstractindextext{%
\begin{abstract}
Effective modeling of human interactions is of utmost importance when forecasting behaviors such as future trajectories. Each individual, with its motion, influences surrounding agents since everyone obeys to social non-written rules such as collision avoidance or group following.
In this paper we model such interactions, which constantly evolve through time, by looking at the problem from an algorithmic point of view, i.e. as a data manipulation task.
We present a neural network based on an end-to-end trainable working memory, which acts as an external storage where information about each agent can be continuously written, updated and recalled.
We show that our method is capable of learning explainable cause-effect relationships between motions of different agents, obtaining state-of-the-art results on multiple trajectory forecasting datasets.
\end{abstract}

}

\maketitle

\IEEEdisplaynontitleabstractindextext

%
\IEEEpeerreviewmaketitle

\section{Introduction}

Autonomous vehicles will soon become a pervasive technology. To comply with safety standards, said vehicles must be able to anticipate what will happen in the surrounding environment. 
Humans perform the same kind of operation when moving in social contexts. In fact, their motion is heavily influenced by group dynamics such as \textit{leader following} or \textit{collision avoidance}~\cite{kothari2020human}. They gather information about nearby entities and their actions are planned according to an estimate of how such entities will behave and move.
This kind of reasoning is possible thanks to a cognitive system known as working memory~\cite{miyake1999models}, which retains short term information and manipulates it in order to make decisions.

To emulate the human capability of forecasting in a dynamic environment, several works have studied mechanisms to model social interactions, in particular with the goal of predicting human trajectories~\cite{helbing1995social, pellegrini2009you, alahi2016social,gupta2018social}.
Recent methods have proposed different solutions, focusing on pooling strategies to aggregate information about individuals~\cite{alahi2016social, kothari2020human, ivanovic2019trajectron}, intra-agent attention mechanisms~\cite{sadeghian2019sophie} or graph-based representations to model both spatial and temporal relationships~\cite{mohamed2020social}.
They have different drawbacks. The main issue of relying on pooling alone is information loss; when pooling multiple signals into a single aggregate, temporal ordering is lost and agent-wise knowledge is disregarded in favor of a coarse global descriptor, which is more practical to handle yet less informative. On the other hand, attention and graph based approaches are able to model relationships between signals but still rely on some fixed-size hidden representation blending them together.
All in all, none of these methods is able to imitate a working memory, which would allow to manipulate individual pieces of information.
Lacking the capability to keep track of each agent, while modeling the overall social interactions, makes it impossible for forecasting methods to properly reason about cause-effect relationships between the motion of different agents.  

\begin{figure}[t]
	\centering
	\includegraphics[width=\columnwidth]{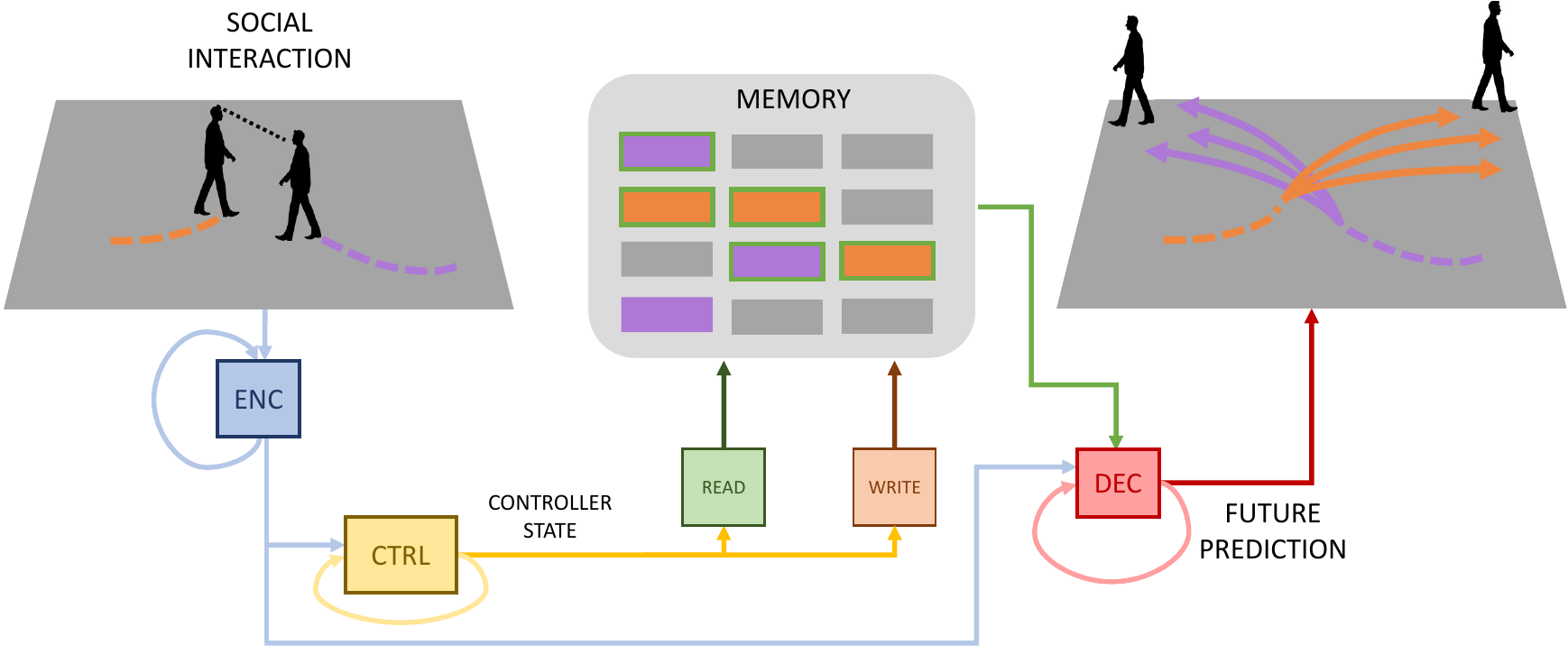}
	\caption{SMEMO models social interaction in trajectory prediction exploiting a working memory. Past observations are encoded and stored in memory and retrieved to formulate multiple  future predictions for each agent.  Read/Write operations are guided by a controller. The whole model is trained end-to-end.}
	\label{fig:eyecatcher}
\end{figure}

Nonetheless, there have been learning based attempts to replicate the cognitive mechanism of a working memory with the definition of Memory Augmented Neural Network (MANN) models~\cite{graves2014neural, weston2014memory, graves2016hybrid}. Differently from previous memory networks, such as RNNs and LSTMs, MANNs add a trainable external memory which is element-wise addressable, i.e. relevant items of information can be accessed selectively. The whole model is end-to-end differentiable, meaning that it learns to store and retrieve relevant information from the memory in a task-driven fashion. State-to-state transitions are obtained through read/write operations.  In this way, MANNs learn to tackle problems in an algorithmic way, i.e., by actively manipulating data to produce an output.  
Successful examples of MANN applications have been proposed for sorting long sequences of data~\cite{graves2014neural}, assigning labels to patterns observed with time-lags~\cite{santoro2016meta} and trajectory prediction of individual vehicles~\cite{marchetti2020memnet, marchetti2020multiple} among others.

We claim that a trainable external memory as provided by the MANN model is utterly important to perform effective reasoning about social dynamics in multi-agent forecasting scenarios, such as human trajectory forecasting in crowds. In fact, it would allow to store and share individual items of information - the positions of the moving agents - and take them into consideration to make predictions at each timestep according to their relevance, without any predefined modeling of their relationship. 

In this paper, we present a model with a Social MEmory MOdule (SMEMO) for trajectory forecasting, where the core element consists of a Memory Augmented Neural Network that is trained to store past information from multiple agents in a shared  external memory and recall and manipulate this information to make predictions. As a byproduct, explicitly storing and reading the trajectory states of the individual agents makes our model interpretable. This allows us to explain cause-effect relationships in our predictions, i.e. to show how a generated trajectory is influenced by the motion of the surrounding agents. Providing explainable predictions is a particularly sensitive aspect when dealing with safety-critical systems, as demonstrated by the increasing interest that this topic has recently started to gain~\cite{kothari2020human, bach2015pixel, selvaraju2017grad}.
We provide evidence of the effectiveness of our solution with application to pedestrian trajectory prediction in crowded environments, where modeling of interactions between the agents is particularly challenging due to  unpredictable rules that govern crowds and the unconstrained variability of agents' movements.


The main contributions of our work are the following:
\begin{itemize}
	\item We propose a new model with an end-to-end external working memory, which we refer to as Social MEmory MOdule (SMEMO), capable of modeling agent interactions for trajectory prediction.
	
	\item Thanks to multiple memory heads, SMEMO is able to generate multiple diverse predictions, addressing the multimodal nature of the task.
	
	\item We show that our model is able to understand social rules in crowded scenarios. We present a new synthetic dataset of social interactions between multiple agents, and show that our solution outperforms the current state of the art models especially in the presence of tight social interaction.

	\item As a direct consequence of explicitly modeling cause-effect relationships between agents' behaviors, SMEMO provides explainable predictions without requiring external tools for interpreting its decision process.

\end{itemize}

\section{Related Work}

In this section we first provide an overview of the state of the art regarding trajectory prediction, with a particular attention on methods that can also provide explainable decisions. We then report a summary of most prominent methods that, as ours, work with memory based models, either as an internal state (RNNs) or as an external module (MANNs).

\textbf{Trajectory Prediction}
Trajectory prediction aims at estimating future locations of moving agents. Difficulties arise from several sources: past motion has to be understood in order to accurately predict the future~\cite{alahi2016social, giuliari2020transformer}; a representation of the environment should be obtained in order to provide spatial cues or constraints on admissible areas~\cite{lee2017desire, srikanth2019infer, chang2019argoverse, caesar2020nuscenes, shafiee2021introvert}; other agents must be taken into account to detect social patterns such as group dynamics~\cite{helbing1995social, pellegrini2009you, alahi2016social, gupta2018social, sadeghian2019sophie, ivanovic2019trajectron, ma2019trafficpredict, lee2017desire, yuan2021agentformer}. \minorrevision{
Several works exploited generative models either to obtain multiple futures forecasts from a single past or to estimate prediction uncertainty~\cite{lee2017desire, gupta2018social,salzmann2020trajectron++,tang2019multiple, mao2023leapfrog, gu2022stochastic}.}

Whereas a fine modeling of motion is always necessary, environment and social behaviors gain importance depending on the scenario and the type of observed agent.
Complex environment representations are useful in automotive settings, where the focus is on road and lane layouts~\cite{chang2019argoverse, caesar2020nuscenes, deo2018multi}.
This is motivated by the fact that vehicles are severely constrained, both in terms of urban regulations and physical ability to roam, since they can accumulate a lot of momentum.
Modeling social patterns often offers a better comprehension of future motion, despite there being evidence of limited relevance with motorized agents~\cite{lee2017desire, chang2019argoverse}. Most prior work has in fact focused on modeling interactions between moving agents while observing pedestrians~\cite{helbing1995social, pellegrini2009you, alahi2016social, gupta2018social, lisotto2019social, ivanovic2019trajectron, salzmann2020trajectron++}.



One of the first works to address social interaction with neural trajectory predictors has been Social-LSTM~\cite{alahi2016social}, which adopted a grid based social pooling to aggregate information about spatially close agents. A follow-up work has extended this idea by adding an adversarial loss to generate socially acceptable multiple future trajectories~\cite{gupta2018social}. The idea of combining recurrent networks to model pedestrian dynamics with an adversarial discriminator has also been followed by~\cite{sadeghian2019sophie}, adding a physical and social attention to focus on interactions. Recently, Shafiee \etal proposed a model integrating a kinematic trajectory representation based on LSTMs with a 3D attention model extracted directly from video sequences, instead of modeling trajectories alone~\cite{shafiee2021introvert}. This approach has the advantage of removing the need of multiple target tracking thus increasing the inference efficiency.

A recent trend followed by several works is to estimate intentions, i.e. generate a likely spatial goal that the agent wants to reach~\cite{Zhao2020TNTTT, Dendorfer_2020_ACCV, mangalam2020not, he2021where, mangalam2021goals}. This kind of technique has proven extremely effective, retaining the current state of the art in several benchmarks. However, solely estimating trajectory goals neglects the social context which can deeply affect agent trajectories. In this work, we do not model intentions since social cues might modify the trajectory of an agent even after the observation horizon, thus bringing it far from the estimated goal in the predicted future.
Furthermore, we highlight the difficulties that intention-based models encounter in highly social environments experimenting on a synthetic dataset where predictions must consider social interaction rules.
Finally, an effective way of modeling social interactions, which is gaining an increasing attention, is to model agents as nodes in a graph and then process it with a Graph Neural Network (GNN)~\cite{mohamed2020social, Kosaraju2019BIGAT, shi2021sgcn}. 
Differently from these methods we model social dynamics by exploiting a trainable external working memory. In this way, we are able to avoid a processing flow that eventually blends past information together into a single latent state, but instead we let the memory able to keep track of relevant cues across time, and store them separately to be successively recalled. To the best of our knowledge we are the first to adopt an external working memory trained end-to-end to model social behaviors and perform trajectory prediction.

A relevant consequence of using an external memory is explaining the decision process of the network. We show that thanks to the Memory Module we are able to highlight cause-effect relationships between agent behaviors, without the usage of external tools to generate interpretable decisions~\cite{kothari2020human, bach2015pixel, selvaraju2017grad}.
A few prior works have underlined the importance of interpretable predictions for trajectory forecasting~\cite{kothari2021interpretable, yuan2021agentformer, kothari2020human, Huang2019stgat}. In~\cite{kothari2021interpretable}, the authors specifically design a model to be interpretable by generating a probability distribution over a discrete set of possible intents, intended as combinations of direction and speed. Such intents however are modeled over four hard-wired social rules.
The most similar work to ours in terms of explainability is~\cite{yuan2021agentformer}, which exploits a form of attention that can attend to information about other agents individually. Such attention is derived from the internals of a transformer architecture. Differently, we rely on read/write weights that our model produces to access memory and attend information from different agents.



\begin{figure}[t]
	\centering
	\includegraphics[width=0.7\columnwidth, trim= 0 50 0 0 clip]{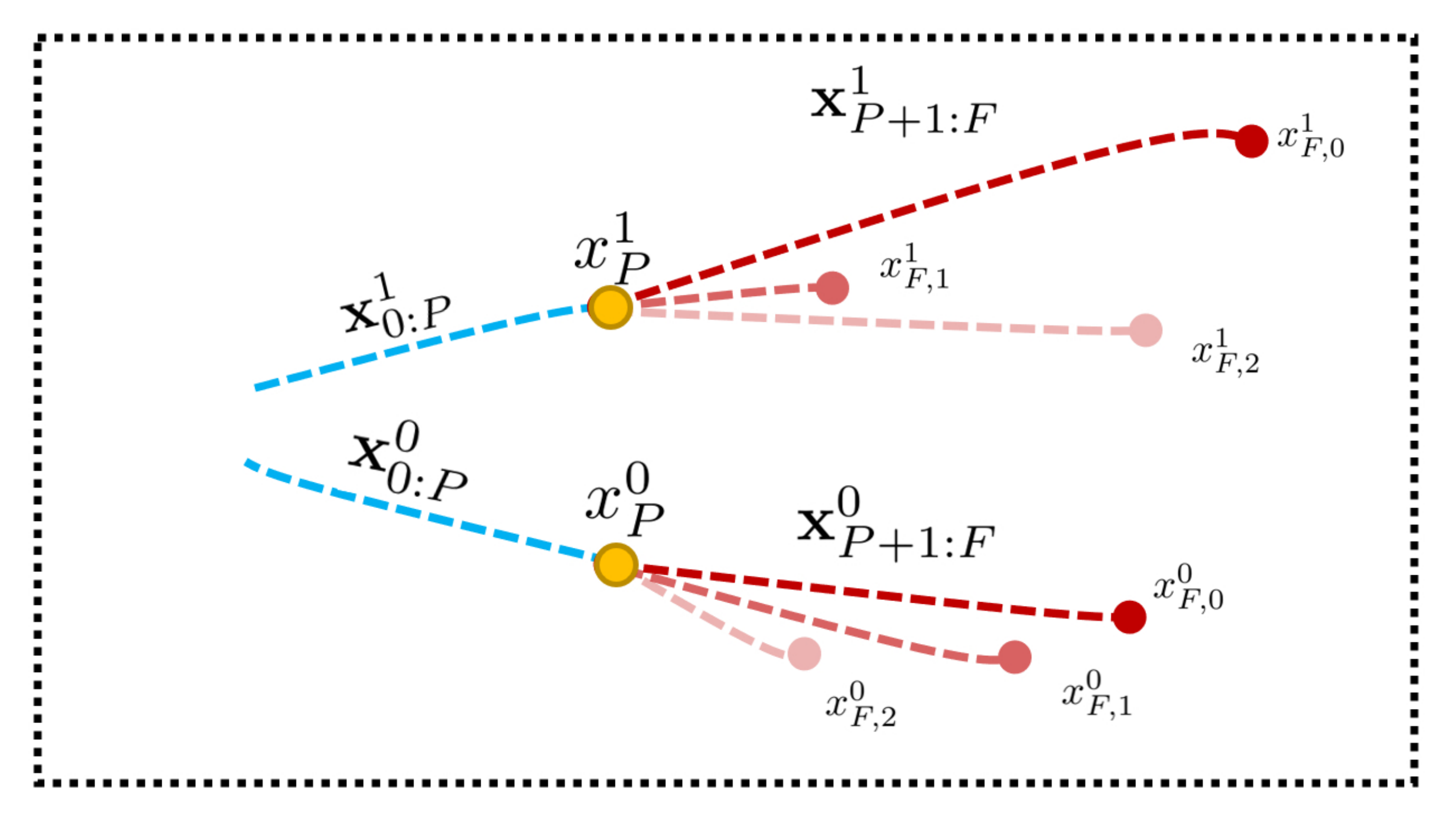}
	\caption{\label{fig:problem_formulation} A social context $\mathcal{S}$ with two agents $\mathbf{x}^0$ and $\mathbf{x}^1$. Past (blue) is observed and multiple futures are predicted (shades of red).} 
\end{figure}

\textbf{Memory Augmented Neural Networks}
Neural Networks are trained to solve specific tasks in a data driven fashion. The whole knowledge is stored in the weights of the model, which learns how to produce adequate responses depending on the input. Tasks involving data sequences, e.g. time series, require to remember early information and hold on to that information up to a point when it might become relevant to produce the answer. Recurrent Neural Networks (RNN) addressed this issue by updating an internal state that attempts to summarize the whole history of observed inputs. Whereas this approach has proven effective in a large variety of cases, it has been shown to suffer from long term forgetting due to exploding/vanishing gradients that affect memorization capabilities. This drawback has been mitigated to some extent by improved versions of RNNs, such as Long Short Term Memories (LSTM)~\cite{hochreiter1997long} or Gated Recurrent Units (GRU)~\cite{cho2014learning}.
RNNs, however, still represent their memory as a latent fixed-size state that will eventually lose track of some information for sufficiently long input sequences. Another limitation of relying on a latent memory is that individual pieces of information cannot be recalled, making it hard to perform reasoning tasks that involve data manipulation. Apparently simple tasks such as copying or sorting become therefore extremely challenging to address.

Memory Augmented Neural Networks (MANN)~\cite{graves2014neural, weston2014memory} are Neural Networks that behave as RNNs in the sense that can be updated through time, but instead of relying on an internal latent state, they exploit an external addressable memory. Such memory is fully differentiable and the model, thanks to a trainable controller, learns to read and write relevant information. The first embodiment of a MANN has been Neural Turing Machine (NTM)~\cite{graves2014neural}, introduced to solve simple algorithmic tasks, demonstrating large improvements when compared to RNNs. Using an external memory, in fact, allows the network to store knowledge that cannot be forgotten unless deleted by the model itself. At each timestep the network can perform reasoning involving all previous observations and can perform data manipulation to emit its outputs.
Follow-up works have extended and refined the formulation of the NTM~\cite{santoro2016meta, sukhbaatar2015end, weston2014memory, graves2016hybrid}. Recently, several declinations of MANNs have been proposed to tackle more complex problems such as online learning~\cite{rebuffi2017icarl}, object tracking~\cite{yang2018learning, lai2020mast}, visual question answering~\cite{kumar2016ask, ma2018visual}, person re-identification~\cite{pernici2020self}, action recognition~\cite{han2020memory} and garment recommendation~\cite{DBLP:conf/icpr/DivitiisBBB20}.

Recently MANTRA, a fist attempt to use Memory Augmented Neural Networks for trajectory prediction, has been proposed by Marchetti \etal~\cite{marchetti2020memnet, marchetti2020multiple}. However, this approach is not end-to-end since each component is trained independently. Moreover, the external memory is a persistent memory populated during training to describe possible future trajectories and perform multimodal predictions.
\minorrevision{The approach has been extended in~\cite{marchetti2022explainable} and~\cite{xu2022remember}. The former (ESA~\cite{marchetti2022explainable}) adds a sparse transformer-based memory controller to improve the effectiveness and the interpretability of the model, whereas the latter (MemoNet~\cite{xu2022remember}) writes endpoint goals rather than encodings of whole trajectories.
Differently from MANTRA and its variants, which completely discards any social component, we exploit an end-to-end trainable episodic memory to reason about social interactions between multiple agents.}

\minorrevision{A different take on using memories for trajectory prediction has been explored in~\cite{yang2022continual}, where a storage is leveraged to address a continual learning setting. Also in this case, the memory is a persistent storage, trained to store samples rather than a working memory as in SMEMO.}

\revision{A few works have used a collection of end-to-end trainable memory cells for trajectory prediction \cite{fernando2018tree, yu2020spatio, li2022graph}, adapting existing data structures to hold temporal information.
In \cite{fernando2018tree}, individual trajectories are fed to a Tree Memory Network (TMN), i.e. a recursive structure where hidden states of an RNN are organized in a hierarchical binary tree, allowing to learn short and long term temporal dependencies.
Differently from \cite{fernando2018tree}, we exploit the external memory as a shared workspace capturing social dynamics between different agents and its read heads to predict multiple futures.
In addition, our memory allows individual agents to perform independent write operations, accessing memory portions selectively, which is not possible with TMN.

In \cite{yu2020spatio} and \cite{li2022graph} instead a graph memory keeps track of the state of each agent at every timestep to perform trajectory smoothing. In \cite{yu2020spatio} memory operations are done without training by appending hidden states. In \cite{li2022graph}, instead, trajectory smoothing is performed by combining a transformer with memory replay. This is an algorithm based on a graph-structured memory storing previous outputs and informing the transformer decoder of previous motion patterns, avoiding sharp turns in the predictions.
The purpose of the memories in both \cite{yu2020spatio} and \cite{li2022graph} is therefore different since memory in our work is a shared working memory that informs every agent about the social context to comply with complex social behaviors.
Memory structure is also different since SMEMO does not rely on a graph but on a memory bank, selectively readable/writable with information of individual agents.}

%
%
%
%

\begin{figure*}[ht]
	\centering
	\includegraphics[width=0.75\textwidth]{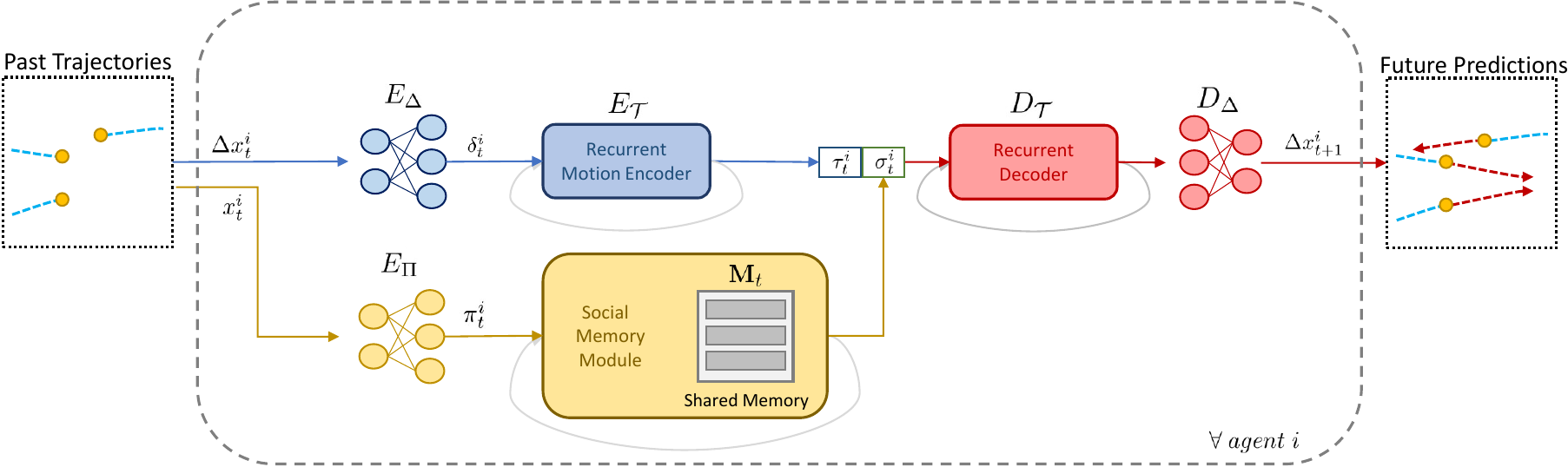}
	\caption{\label{fig:architecture}Past trajectories are fed to the Egocentric stream (Blue) and a Social stream (Yellow). In the Egocentric stream, each past trajectory is embedded and then encoded with a Recurrent Motion Encoder GRU. In the Social stream, the Social Memory Module encodes past trajectories and populates the working memory, which is shared for each agent. This allows the model to create a representation for the social context. Prediction is performed by decoding the concatenation of the past encoding and the social feature.} 
\end{figure*}

\section{Method}


\subsection{Problem Formulation}
\label{sec:problem_formulation}

Given a social context $\mathcal{S}=\{\mathbf{x}^i, i=0,...,N-1 \}$, defined as the set of trajectories representing $N$ moving agents, we formulate the task of trajectory prediction as the problem of predicting the future positions of each agent, given their past positions. 
We consider sequences of agent trajectories belonging to different social contexts as independent episodes. Within each episode, trajectories span from an initial observation point up to a prediction horizon.




All trajectories, past and future, are sequences of top-view 2D spatial coordinates in a fixed reference frame, independent from the agents. Past trajectories are observed over a temporal interval up to an instant $P$, identified as the present $ \mathbf{x}_{0:P}^i\ = \{x_0^i, x_1^i, ..., x_{P}^i \}$. All movements taking place after the present and up to an instant $F$ belong to future trajectories $\mathbf{x}_{P+1:F}^i \ = \{x_{P+1}^i, ..., x_{F}^i \}$.

Trajectory prediction is an inherently multimodal problem, meaning that given a single observation, multiple outcomes are possible. Following the recent literature \cite{gupta2018social, lee2017desire, salzmann2020trajectron++} we generate multiple future estimates to provide a variety of futures in order to cover this uncertainty.
To provide a comprehensive notation, throughout the paper we define variables with a superscript identifying the referred agent and two subscripts indicating the current timestep and current future estimate identifier. Therefore the $t$-th timestep of the $k$-th future prediction for agent $i$ is denoted as $x_{t,k}^i$.
Fig.~\ref{fig:problem_formulation} exemplifies such notation, depicting a social context with two agents and three diverse future predictions.
	

\subsection{Architecture Overview}
\label{sec:overview}


In SMEMO, the motion of each agent is processed into two streams, which we refer to as Egocentric and Social, as shown in Fig.~\ref{fig:architecture}. The former is dedicated to modeling relative displacements of an agent from one timestep to another. This allows to understand how individual agents move, regardless of their actual position in space. The latter instead, processes the absolute agent positions to obtain knowledge of where an agent is with respect to the environment. This information is then stored into an external memory, shared across agents. Our model therefore can learn to perform social reasoning by manipulating memory entries to predict future positions for all agents in the scene.

In the Egocentric Stream, at each timestep $t$, past displacements $\Delta x^i_t$ are observed for each agent trajectory $\mathbf{x}^i \in \mathcal{S}$. Each displacement is first processed by an encoder $E_{\Delta}$ to obtain a projection $\delta^i_t$ into a higher dimensional space. The temporal sequence of $\delta^i_t$ is then fed to a recurrent motion encoder $E_\Tau$, which generates a condensed feature representation $\tau^i_{t}$.

In the Social Stream, past absolute positions $x_t^i$ are considered for each agent trajectory $\mathbf{x}^i \in \mathcal{S}$. A projection $\pi_t^i$ is obtained with an encoder $E_{\Pi}$. This yields a sequence of temporized descriptors, which is directly fed to the Social Memory Module. 
This module acts as a recurrent neural network and processes a sequence of input features in parallel for each agent. It generates a compact social descriptor $\sigma^{i}_{t}$, summarizing social behaviors between all agents in the social context $\mathcal{S}$ up to the current timestep $t$.
The $i$ superscript denotes a separate social descriptor for each agent, beyond the fact that all participate in a common social context. This is necessary since agents interact differently with the others depending on their position and movement.

The egocentric and social representations, $\tau^i_{t}$ and $\sigma^i_{t}$, are finally concatenated and fed to a recurrent motion decoder $D_\Tau$ and the model autoregressively predicts future displacements $\Delta x^i_{t+1}$, for each agent,  with a decoder $D_{\Delta}$.
Each autoregressive step works as follows.
$E_\Delta$ and $E_{\Pi}$ respectively process each $\mathbf{\Delta x}^i_{0:P}$ and $\mathbf{x}_{0:P}^i$ independently, generating at each timestep the latent representations $\delta^i_t$ and $\pi^i_t$, until the present is reached.
For each timestep in the future, instead, $\delta^i_t$ and $\pi_t^i$ are replaced with a vector of zeros to allow the autoregressive trajectory generation. The recurrent encoder $E_\Tau$ and the Social Memory Module therefore keep updating their internal state and new $\tau^i_{t}$ and $\sigma^i_{t}$ are generated for each instant in the future.
All agents share $E_\Delta$, $E_\Tau$, $E_\Pi$, $D_\Tau$ and $D_\Delta$ and the memory $\textbf{M}$ of the Social Memory Module. 

\subsection{Social Memory Module}
\label{sec:memory}
\revision{The Social Memory Module serves the purpose of reasoning about interactions between all the observed agents.
The module is composed of an external memory $\textbf{M} = [\textbf{m}_0, \textbf{m}_1, ..., \textbf{m}_{|\textbf{M}|-1}]$, shared across all agents.
Each memory entry $\textbf{m}_j$ is a readable/writable cell of dimension $Q$. Overall, \textbf{M} can be interpreted as a matrix with dimensions $|\textbf{M}| \times Q$.
The Social Memory Module interacts with the memory using a recurrent controller with a single write head and multiple read heads.
As in standard Memory Augmented Neural Networks (MANN)~\cite{graves2014neural, weston2014memory, sukhbaatar2015end}, $\textbf{M}$ is an episodic memory used as workspace for writing and reading relevant information from a data sequence, which is then wiped out at the end of each episode.}


The Social Memory Module aggregates data from all agents in the social context $\mathcal{S}$ and continuously outputs a social feature $\sigma^i_{t}$, which condenses the history of the whole episode up to the current timestep $t$.
The module is fed with multiple sequences of encoded absolute positions $\pi_t^i$ of $i=0,...,N-1$. 
For each timestep $t$, the encoded absolute positions of each agent are given in input in parallel.
When each element of a sequence $\pi_t^i$ is presented to the module, it is concatenated with the social feature generated at the previous step by the Social Memory Module itself. Such social feature is initialized with a zero-vector and updated autoregressively at each timestep. The obtained representation is then fed to the Memory Controller, which outputs a latent feature $\gamma^{i}_{t}$ used to access memory through read and write operations.
These operations happen subsequently at each timestep: first the module accesses memory to fetch relevant social information (read phase), then the memory is updated to take into account the current observation (write phase).
\revision{Since $\textbf{M}$ is continuously updated, we denote with $\textbf{M}_t$ the memory content at timestep $t$.}


\begin{figure}[t]
	\centering
	\includegraphics[width=\linewidth]{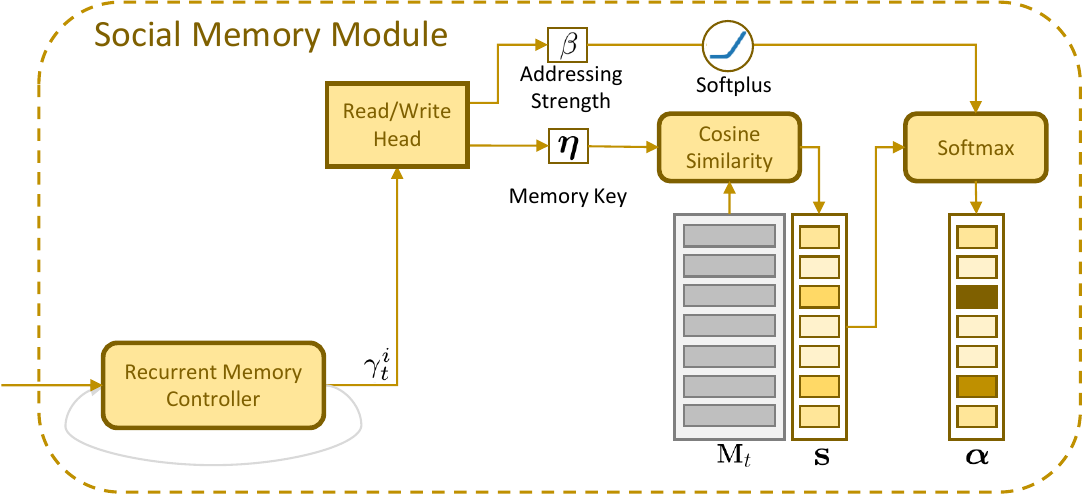}
	\caption{\label{fig:addressing} Social Memory Module Addressing. SMEMO is equipped with a shared memory $\textbf{M}_t$. The controller outputs at each timestep a feature $\gamma^{i}_{t}$ which is fed to the read/write heads to generate a memory key $\boldsymbol{\eta}$ and an addressing strength $\beta$. The key is used to find relevant memory locations in memory via cosine similarity. Access weights $\boldsymbol{\alpha}$ are then obtained by normalizing such similarities through a softmax with temperature $\beta$.} 
\end{figure}

\begin{figure}[t]
	\centering
	\includegraphics[width=\linewidth]{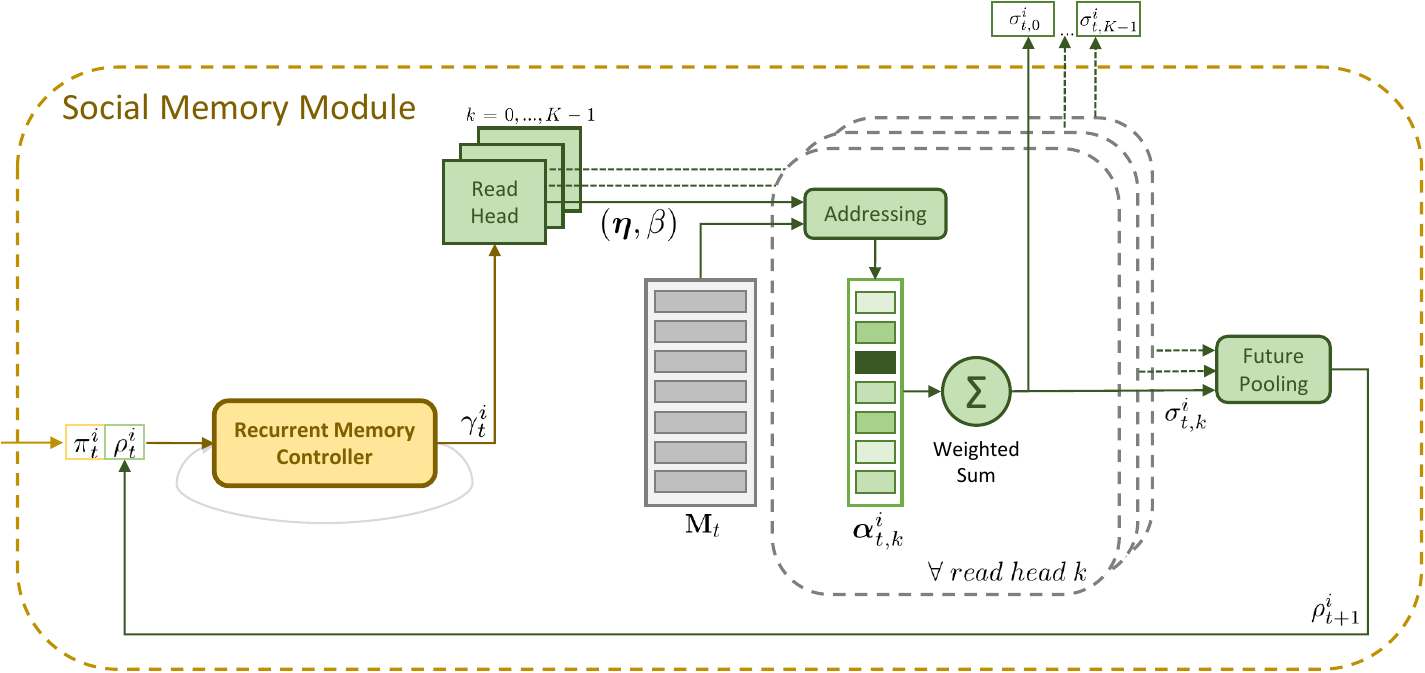}
	\caption{\label{fig:reading}Social Memory Module Reading. For each agent $i$, separate read heads perform a memory addressing to obtain $K$ social features $\sigma^i_{t,k}$ which will be fed in parallel into the decoder to generate a multimodal future prediction. The social features are then pooled together via Future Pooling and fed back to the model auto-regressively.} 
\end{figure}

\subsubsection{Addressing}
\label{sec:addressing}
Read and write operations share a memory addressing step, where the most relevant memory locations are identified by generating read/write weights $\boldsymbol{\alpha}$ (Fig. \ref{fig:addressing}). In this section we refer to these weights as access weights, regardless if they are used for reading or writing, even if they are exploited in different ways as outlined in Sec.~\ref{sec:reading} and Sec.~\ref{sec:writing}. For the sake or simplicity, in this section we temporarily drop the agent superscript and timestep subscript, reminding the reader that addressing operations are executed for all the agents at each timestep.

Memory addressing is performed by read/write heads starting from the state of the memory controller $\gamma$. Each head is a dense layer that transforms the state into a pair $(\boldsymbol{\eta}, \beta)$, which respectively indicate the memory key and addressing strength. The weights generating $\beta$ have a softplus activation, as in Neural Turing Machines \cite{graves2014neural}, to ensure it to be grater than zero.

The memory key $\boldsymbol{\eta}$ is used to find relevant memory locations $\textbf{m}_j$ via cosine similarity $s_j = \frac{\boldsymbol{\eta} \cdot \textbf{m}_j}{\| \boldsymbol{\eta} \| \cdot \| \textbf{m}_j\|}$. To obtain the final access weights, $\beta$ is used to control the amount of focus by normalizing the similarities $\mathbf{s}=\{s_0, ,..., s_{|\textbf{M}|-1}\}$ with a learned temperature-softmax, where $|\textbf{M}|$ is the number of memory entries:


\begin{equation}
\alpha_j = \frac{   e^{\beta s_j}      }{
	\sum_l e^{\beta s_l}
} ~~~~~~~ j=0,...,|\textbf{M}|-1
\end{equation}




\subsubsection{Reading}
\label{sec:reading}
The Social Memory Module is equipped with $K$ read heads. These heads are used in parallel to produce a multimodal prediction, i.e., a variety of futures.
Each read head performs a memory addressing operation (see Sec. \ref{sec:addressing}), obtaining read weights $\boldsymbol{\alpha}_{t,k}^{i}, k=0,...,K-1$.

Assuming that the memory contains relevant information concerning social dynamics of the agents, a social feature $\sigma^i_{t,k}$ can be produced by simply performing a sum over memory locations, weighted by $\boldsymbol{\alpha}_{t,k}^{i}$. This feature is then directly passed to the Egocentric Stream and decoded into a multimodal prediction by applying the decoder for each future $k$ in parallel. Having $K$ read heads allows the model to learn different modalities and to condition the decoder to generate $K$ diverse futures.
Finally, an autoregressive social feature $\rho_{t+1}^i$, to be fed to the model at the next timestep, is obtained by applying a max-pooling operation to $\sigma^i_{t,k}$ over all $k=0,...,K-1$ futures. We refer to this step as \textit{Future Pooling}. Future Pooling is necessary to obtain a compact descriptor summarizing all $K$ futures and avoid the combinatorial growth of agents and futures, which would require a large amount of Memory Controllers evaluated in parallel. The pooled feature $\rho_t^i$ is then fed back as input to the Social Memory Module at the next time-step as shown in Fig. \ref{fig:reading}.

\begin{figure}[t]
	\centering
	\includegraphics[width=\linewidth]{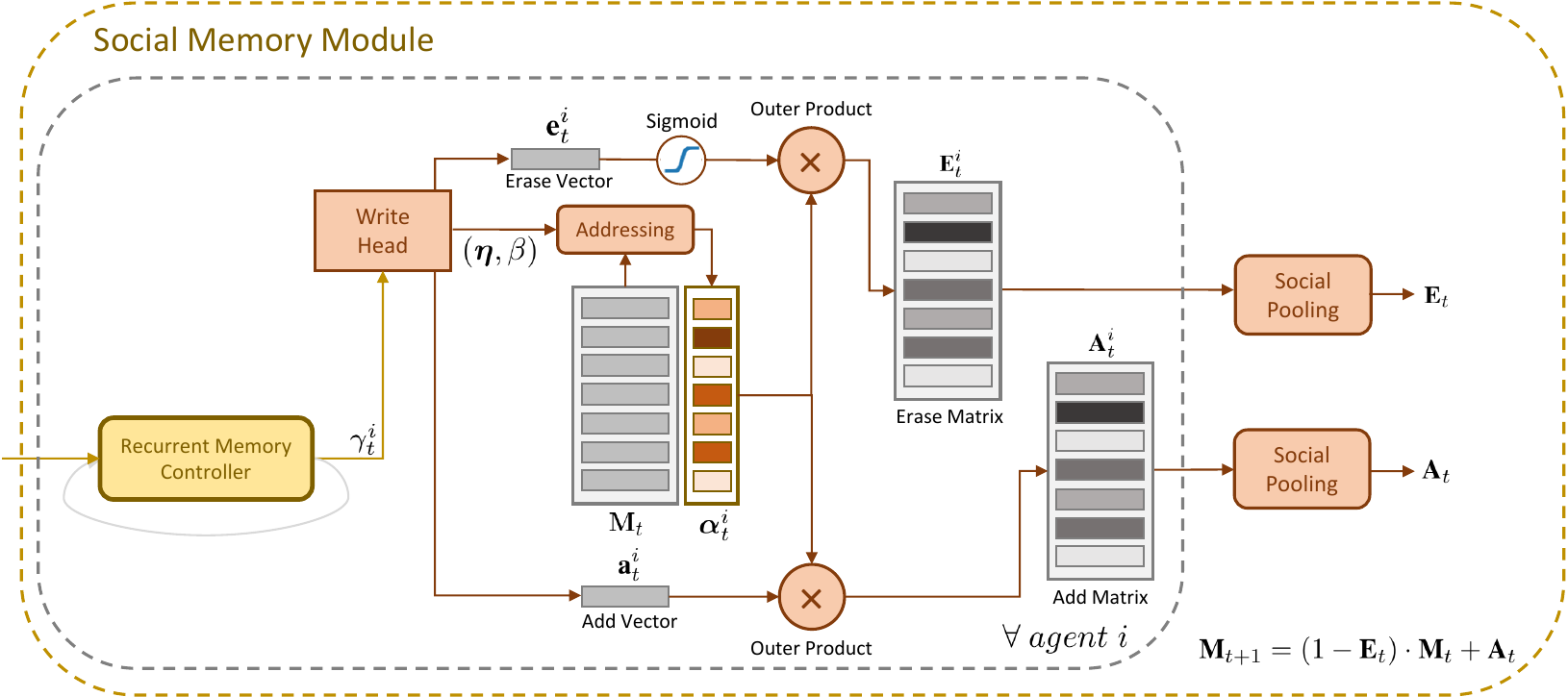}
	\caption{\label{fig:writing}Social Memory Module Writing. For each agent, a single write head is in charge of addressing memory and generating erase and add vectors. The vectors are combined with the write weights to generate erase and add matrices. Memory update is performed after Social Pooling to achieve invariance to the order in which agents write in memory at each timestep.}
\end{figure}

\subsubsection{Writing}
\label{sec:writing}
To store relevant information for each agent $i$, a write head produces write weights $\boldsymbol{\alpha}^i_t$ through memory access (see Sec. \ref{sec:addressing}), an erase vector $\textbf{e}^i_t$ and an add vector $\textbf{a}^i_t$.
\revision{The two vectors are generated by two additional dense layers which, similarly to the generation of the memory key, transform the controller state $\gamma$ into $\textbf{e}^i_t$ and $\textbf{a}^i_t$ respectively.}
The former is passed through a sigmoid activation to bound between 0 and 1 the rate of information that can be erased, i.e. when $\textbf{e}^i_t=0$ no information is erased, instead when $\textbf{e}^i_t=1$ all the information is erased.
The write weights are combined with $\textbf{e}^i_t$ and $\textbf{a}^i_t$ with an outer product to obtain add and erase matrices, which will be used to update each memory entry:
\begin{equation}
\textbf{A}^i_t = \boldsymbol{\alpha}^i_t \otimes \textbf{a}^i_t\quad\quad \textbf{E}^i_t = \boldsymbol{\alpha}^i_t \otimes \textbf{e}^i_t
\end{equation}

Since we want the shared memory to contain information about each moving agent, before updating the memory content, we first generate erase and add matrices $\textbf{E}^i_t$ and $\textbf{A}^i_t$ for each agent $i$ and then perform a max pooling operation $\textbf{E}_t = \max \{\textbf{E}^i_t\},~ \textbf{A}_t = \max \{\textbf{A}_t^i\}$ across all agents $ i=0,...,N-1$. This mechanism is similar in spirit to \textit{Social Pooling}~\cite{alahi2016social, deo2018convolutional}, where states from different agents are pooled together to condition predictions. The need of a social pooling stems also from the fact that we want the memory content to be invariant to the order in which agents are observed at each step. Memory update is performed after each timestep as

\begin{equation}
\label{eq:update}
\textbf{M}_{t+1}= (1-\textbf{E}_t)\cdot\textbf{M}_{t} + \textbf{A}_t.
\end{equation}

where we use the subscripts $t$ and $t+1$ to denote the memory content at two consecutive timesteps.
The memory writing process is shown in Fig.~\ref{fig:writing}.

\begin{figure}[t]
	\centering
	\includegraphics[width=.8\columnwidth]{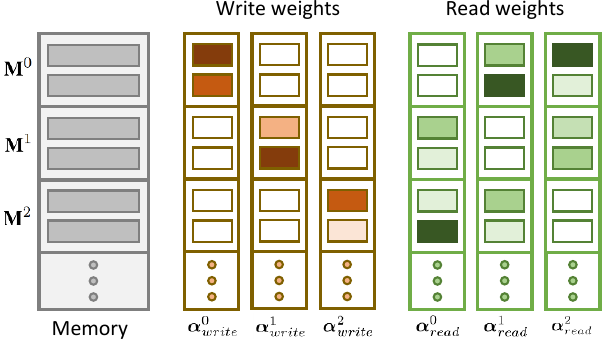}
	\caption{\label{fig:explainability_memory} Memory configuration for explainable predictions. Memory is divided in segments $\textbf{M}^i$, reserved to individual agents. Information for agent $i$ is written only in $\textbf{M}^i$ and read from all the other segments. White-colored addressing weights refer to zeroed-out cells. $\boldsymbol{\alpha}^i_{write}$ and $\boldsymbol{\alpha}^i_{read}$ indicate the writing and reading addressing for agent $i$.
	}
\end{figure}

\section{Explainability}
\label{sec:exp}

In addition to correctly predicting the future trajectory of an agent, it is interesting to understand which neighbors have contributed to the generation of such trajectory.
In particular, understanding how agents are influenced by their neighbors can foster explicit modeling of cause-effect relationships and provide explainable predictions. This aspect is of fundamental importance in safety critical autonomous driving systems. In fact, providing a certain degree of explainability can improve the welcoming of such technology while providing a higher level of understanding about the internals of the system itself.
This problem is not trivial and most trajectory predictors do not provide explainable predictions by design. This is due to the "black box" structure of neural networks which makes an output non interpretable and does not provide any information about \textit{why} a predicted trajectory follows a certain pattern.
The main peculiarity of SMEMO is the use of a shared episodic memory where, at each timestep, agents store information explicitly. 

In the following we show how, with a simple constraint on where the memory controllers can write and read, our model can become easily interpretable by design.
\revision{During the writing phase we reserve a segment of $z$ memory cells for each agent. In this way, we can compartmentalize the information about the social context agent by agent. We refer to $\textbf{M}^i = [\textbf{m}_{i \cdot z}, ..., \textbf{m}_{i \cdot z + z - 1}]$ as the set of $z$ memory entries reserved to agent $i$. Therefore, the overall memory can be re-defined as $\textbf{M}=[\textbf{M}^0, \textbf{M}^1,..., \textbf{M}^{N-1}]$ where $N$ is the number of agents in the social context.  In our experiments we use $z=5$.}
Moreover, to predict the future position of a certain agent, we force SMEMO to read only segments reserved to other agents. As a consequence, memory access during the reading phase can be interpreted as an inter-agent attention, underlining the importance of other agents during each step of the generation of a future trajectory.
Fig.~\ref{fig:explainability_memory} depicts how a controller can read and write in memory in this configuration, depending on the trajectory to be predicted. For the sake of simplicity, we drop the timestep subscript, reminding the reader that writing/reading operations are executed for each timestep.

Thanks to this formulation, cause-effect relationships emerge since it is possible to observe on which memory segment the reading controller focuses its attention before predicting the future of an agent.
To perform such analysis, it is sufficient to examine how features in memory are updated and how the controller is attending them to generate predictions, without requiring further training of the network.
\revision{Referring to Fig. \ref{fig:explainability_memory}, we define the attention for an agent $i$ on another agent $j$ as the sum of all the reading weights of the segment dedicated to $j$, i.e. $\textbf{M}^j$:}


\begin{equation}
\textbf{att}^i(j) = \sum_{c = j \cdot z}^{j \cdot z + z}{\boldsymbol{\alpha}^i_{read}(c)}
\end{equation}

\revision{where $\boldsymbol{\alpha}^i_{read}(c)$ indicates the $c$-th element of the read vector $\boldsymbol{\alpha}_{read}$ generated for agent $i$ and $z$ is the number of cells in the segment $\textbf{M}^j$.}
The attention for agent $i$ on all other agents is thus defined as 
$\textbf{att}^i = \{\textbf{att}^i(j),  j=\{0,...,N-1 \} \setminus i     \}.$


Finally, we normalize the attentions using a softmax to make the vector sum up to 1. Note that we discard self-attention, i.e., we do not consider the attention of an agent on itself. Such attention however will be zero by construction, since the controller can only read in memory segments reserved to agents different from the one it is predicting.

To provide a better intuition, we provide a simple example involving three moving agents, which we identify as $A0$, $A1$ and $A2$.
Suppose that, at a certain timestep, agent $A1$ and agent $A2$ are about to collide. The model will generate a prediction such that they are able to avoid each other.
That to do so, SMEMO changes the trajectory of $A1$ to avoid $A2$. To make this possible, the controller will need to read $A2$'s state from memory and therefore the reading phase for agent $A1$ will mostly attend the memory segment reserved to $A2$. In this case, the attention value for agent $A1$ on agent $A2$ must be higher compared to that on the agent $A0$, which does not interact with $A1$.

We show in Sec. \ref{sec:explainability_res} that when social behaviors emerge in predicted trajectories, it is possible to highlight the importance of surrounding agents for each prediction of the model.

\section{Experiments}
Assessing the actual capability of a model to address social interactions is not straightforward. The main limitation lays in limited or unlabeled samples of social behaviors in standard datasets.
With this in mind, we first showcase the ability of our model to perform high level reasoning on interaction scenarios with a synthetic trajectory forecasting dataset. We built this dataset to contain social episodes between multiple agents which have to follow specific interaction rules. This is an algorithmic task with a well defined output depending on its initial conditions. We show that common approaches do not perform well under these conditions and fail to understand how the movement of an agent influences the others.
The synthetic dataset also underlines the risks of relying on goal-based approaches, which have become a popular choice in recent trajectory forecasting methods \cite{Zhao2020TNTTT, Dendorfer_2020_ACCV, mangalam2020not, he2021where}.
We then experiment on the Stanford Drone (SDD) and UCY/ETH datasets to provide a comparison with the state of the art on real data.
\subsection{Datasets}
\label{sec:ssa}

\textbf{Synthetic Social Agents}
As outlined in the seminal paper by Helbing \textit{et al.}~\cite{helbing1995social}, pedestrians obey implicit social rules following behavioral patterns. These are not caused by deliberate actions but are rather a reaction to social forces that make individuals influence each other.
These interactions in complex environments are not easy to model and, even when performing well in terms of prediction metrics, trajectory forecasting models might fail to grasp the underlying rules that dominate crowds.
\revision{In real datasets, there are a lot of variables at play that contribute to the complexity of the problem, making such rules less evident and harder to model: the movement of an agent influences the trajectory of the others and since each agent can arbitrarily alter its motion, the problem is inherently multimodal, i.e. several predictions are needed to approximate the future. All these elements concur to make it extremely hard if not impossible to precisely assess and annotate cause-effect relationships in motion patterns of agents that conciliate their attempt to reach a destination with such aforementioned rules.}

To measure the capability of a model to perform correct predictions for this kind of patterns, we synthetically generated a dataset of interacting moving agents that obey a few simple social rules. \revision{This dataset is intended to be an oversimplification of real social behaviors, yet offers a challenging shift of attention from individual motion to social forces. We completely remove multimodality, making motion patterns depend solely on the social context.} We refer to it as the \textit{Synthetic Social Agents} (SSA) dataset.

The dataset is divided into short episodes. Episodes portray a set of agents moving in a scene. In each episode, agents start moving from a random point lying on a fixed-radius circumference towards the middle. The goal of each agent is to cross the circle, passing through the center, without hitting the others.
We define a set of arbitrary rules, which we expect forecasting models to learn in order to perform correct predictions. In our case, the only rules that control the social forces of the samples in the dataset are the following:

\begin{itemize}
	\item When an agent is moving, it maintains a constant speed, drawn at random at the beginning of the episode,  throughout the whole episode and moves forward until the episode is over.
	\item If the trajectories of two agents will intersect within a fixed distance radius $r$, the slowest agent stops until the fastest one has passed and the collision will no longer occur. \minorrevision{We fix $r=1.2m$ in our experiments}.
\end{itemize}


\minorrevision{For each episode, a variable number of agents (from 3 to 10) and their velocity are drawn at random between 1m/s and 2m/s to simulate both slow and fast walks. Agent starting points are quantized along the circumference to avoid agents to be too close and the whole circle is rotated by a random degree to increase sample variability. Samples have a temporal extent of 60 timesteps, 20 in the past and 40 in the future. We fix the radius of the circumference to 6 meters, in order to simulate a plausible interaction scenario.
We defined a fixed train and test split by collecting 9000 and 1000 episodes respectively.}

The challenge behind SSA is given by the necessity of jointly manipulating all observations in order to understand when an agent causes another to react and stop.
\revision{The purpose of this dataset is not to leverage synthetic data to train trajectory predictors that work in the real world as done by prior works \cite{berlincioni2020multiple, buhet2019conditional}, but rather to provide a tool for testing the ability to identify social interactions.}
We argue that this capability has not yet been fully achieved by state of the art trajectory forecasting methods. 
We will publicly release the dataset to foster research on this subject and ease comparisons with future works.

\textbf{Stanford Drone Dataset}
The Stanford Drone Dataset \cite{Robicquet2016sdd} (SDD) is a dataset of agents of various types (pedestrians, bicycles, cars, skateboarders) who move around the university campus acquired through a bird's eye view drone at 2.5 Hz.
The train-test split used for the experiments is that of the Trajnet challenge \cite{sadeghian2018trajnet} (also commonly adopted by other state-of-the-art methods), which focuses on the prediction of pedestrian trajectories. There are about 14k scenarios with multiple agents moving in the scene expressed in pixel coordinates.

\textbf{ETH/UCY}
ETH~\cite{pellegrini2010eth} and UCY~\cite{lerner2007ucy} are two datasets of real-world pedestrian trajectories in top-view coordinates in meters. The acquisition was carried out with a fixed camera on 5 different scenarios captured at 2.5 Hz. ETH contains two scenarios (ETH, HOTEL) while UCY contains three (UNIV, ZARA1, ZARA2). In total we have 5 scenarios with 1536 unique pedestrians and with non-trivial social interaction between agents like group actions, collision avoidance and crossing trajectories.
Following the configuration used by state of the art methods, for the evaluation we use the leave-one-out strategy where we train the model on all trajectories of 4 scenarios and test it on the fifth scenario.

\begin{figure}[t]
	\centering
	\resizebox{0.5\textwidth}{!}{
		\begin{tabular}{cccccc}
			\multicolumn{6}{c}{\includegraphics[width=.55\textwidth]{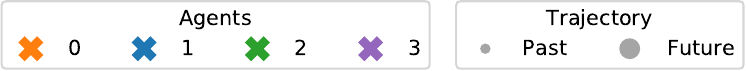}} \\
			
			\multicolumn{2}{c}{\includegraphics[width=.329\textwidth]{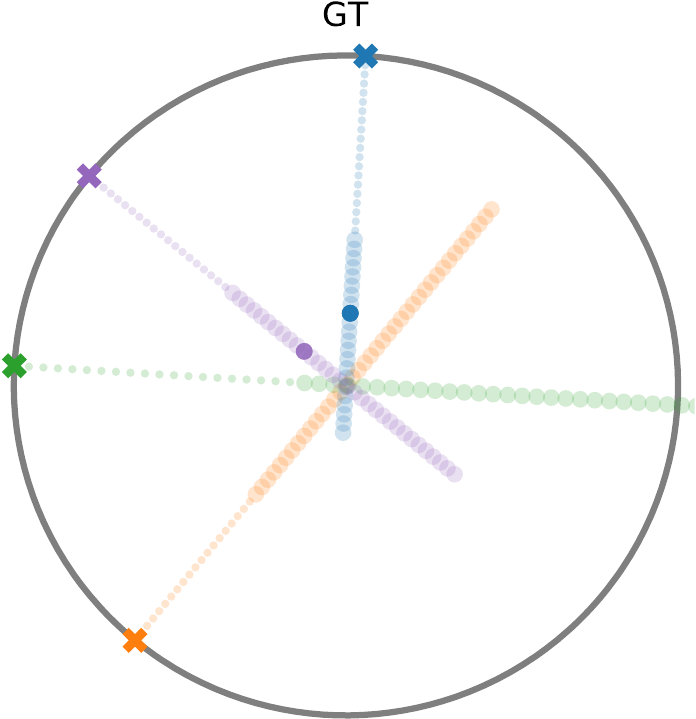}} &
			\multicolumn{2}{c}{\includegraphics[width=.329\textwidth]{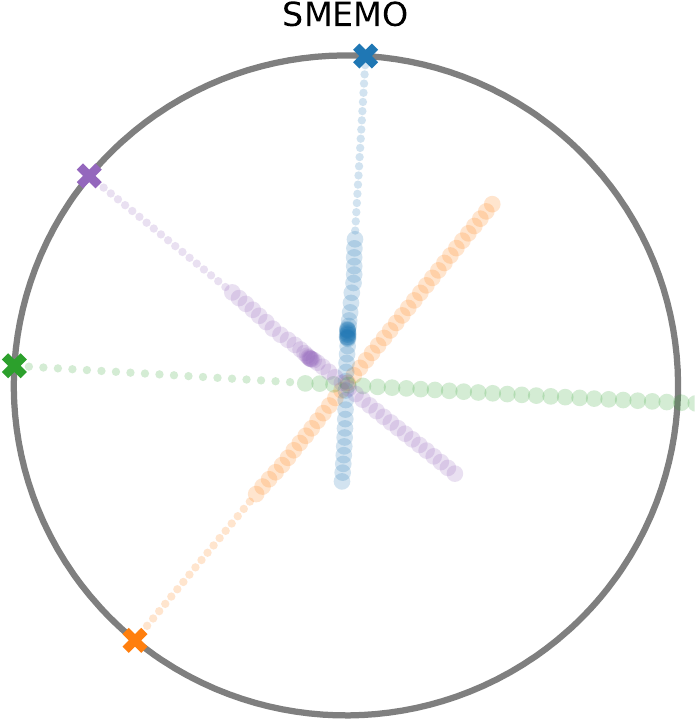}} &
			\multicolumn{2}{c}{\includegraphics[width=.329\textwidth]{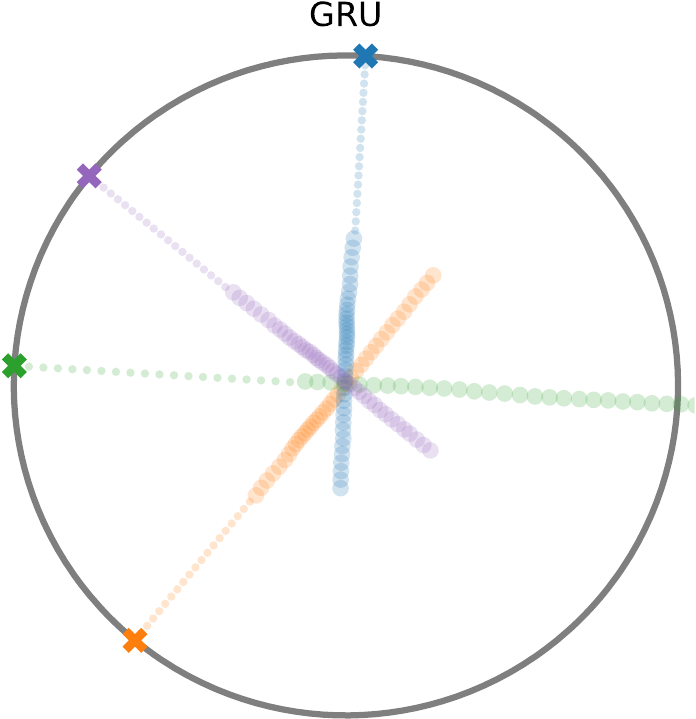}} \\
			
			\multicolumn{3}{c}{\includegraphics[width=.49\textwidth]{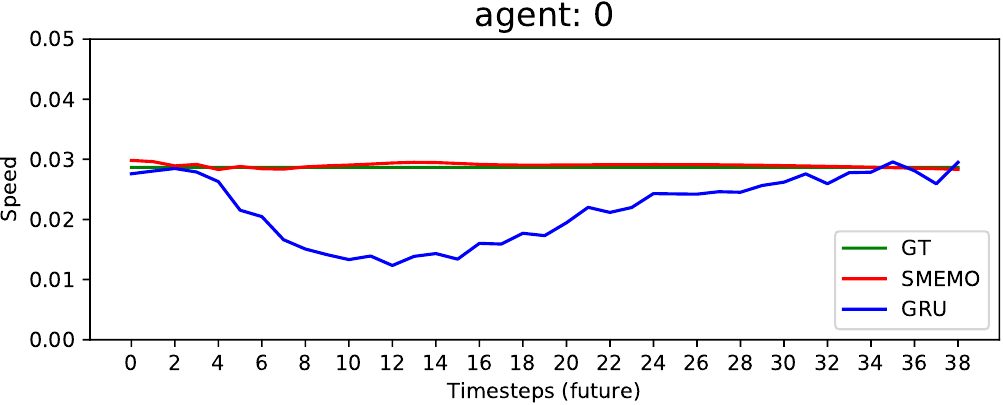}} & \multicolumn{3}{c}{\includegraphics[width=.49\textwidth]{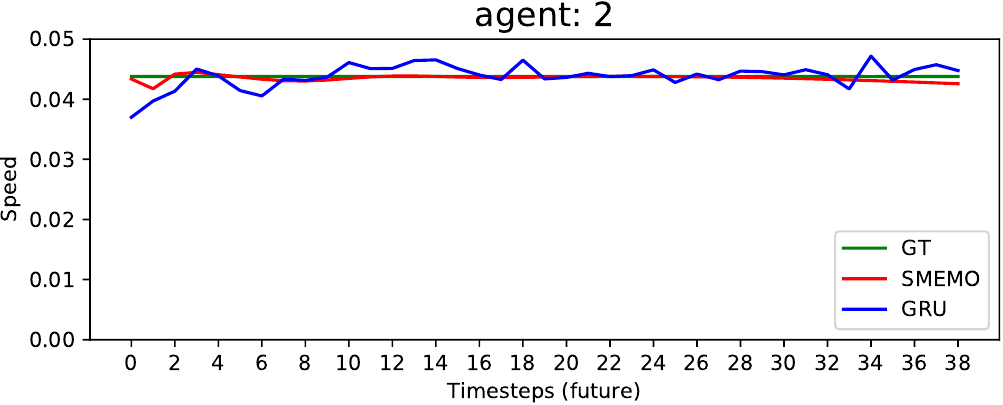}} \\ \multicolumn{3}{c}{\includegraphics[width=.49\textwidth]{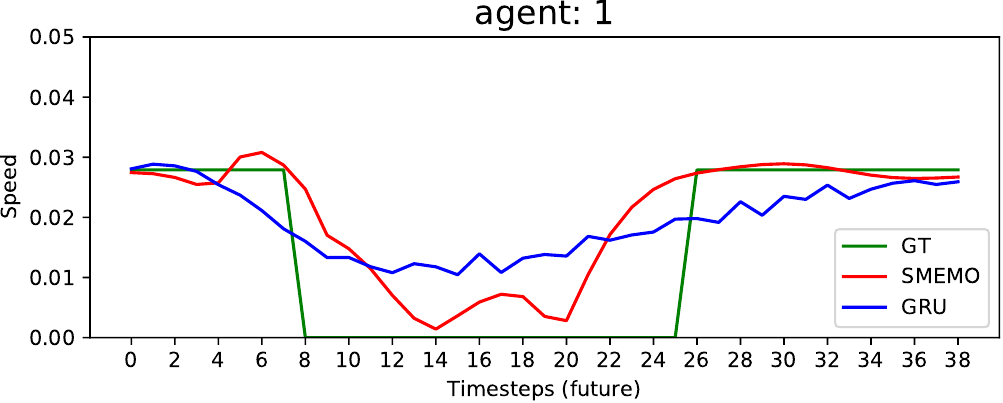}} & \multicolumn{3}{c}{\includegraphics[width=.49\textwidth]{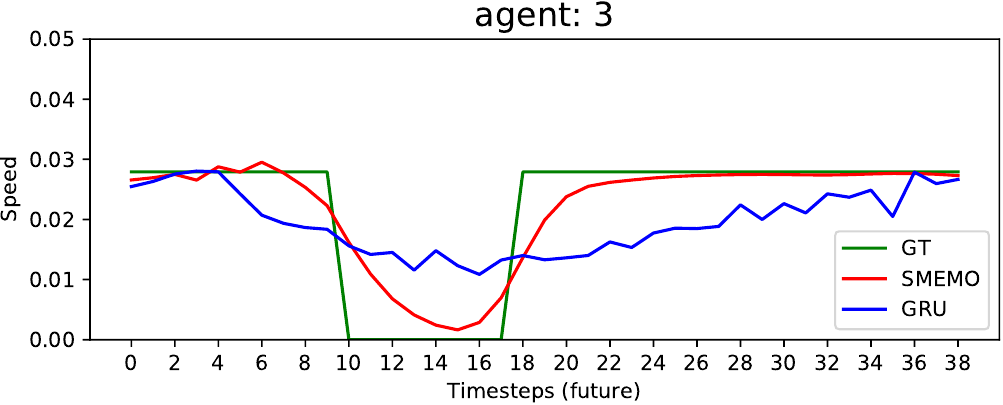}} \\ 
			
		\end{tabular}	
	}
	\caption{
		\label{fig:toy_sample}Comparison of GT with SMEMO and GRU Encoder-Decoder from an episode of SSA. \textit{Top}: agents start from a random point on the circle and approach the center with fixed velocities. Blue and purple agents halt (dark dot in GT), waiting for the orange and green ones to pass. \textit{Bottom}: velocities of each agent during the episode. SMEMO manages to slow down predictions when social interactions occur.}
\end{figure}

\subsection{Training and Implementation}

Following previous literature, for SDD and ETH/UCY we train our model to observe 3.2 seconds of past trajectory and
predict 4.8 seconds in the future for all agents in the scene (respectively 8 and 12 timesteps). In SSA instead we observe 20 timesteps and predict the remaining 40, simulating a 2s past and 4s future sampled at 10Hz. All three datasets are organized in episodes, i.e. several agents are observed simultaneously sharing the same past and future timespans.

The encoders $E_\Delta$ and $E_\Pi$ are Multi-Layer Perceptrons (MLP) composed of two fully-connected layers with a ReLU activation in the middle. The encoders have the purpose of lifting the 2-dimensional signal to a higher dimension, projecting each point to a 16-dimensional representation.
The recurrent motion encoder and memory controller are Gated Recurrent Units (GRU) with a 100-dimensional state. 
The read heads and write head are implemented as fully-connected layers. In particular, to perform memory addressing, they both map the controller state $\gamma^i_{t}$ into the scalar addressing strength $\beta$ plus memory key $\boldsymbol{\eta}$. The write heads instead have two additional fully-connected layers to generate the add and erase vectors $\textbf{a}^i_t$ and $\textbf{e}^i_t$. The three latent variables $\boldsymbol{\eta}$, $\textbf{a}^i_t$ and $\textbf{e}^i_t$ all share the same dimension with the memory entries, which we chose to be Q=20. \revision{Overall, the memory size is fixed at $128 \times 20$, i.e. a memory with states in $\mathbb{R}^{20}$ and $|\textbf{M}|=128$ writable slots.}
Finally, the recurrent motion decoder is a GRU with a 120-dimensional state, followed by a linear layer generating 2D spatial offsets. The dimension of the GRU hidden state stems from the concatenation of the encoded motion $\tau^i_{t}$ and the social feature $\sigma^i_{t,k}$, respectively of size of 100 and 20.

The entire model is trained end-to-end with the Adam optimizer using a learning rate of 0.001 and a batch size of 32 for ETH/UCY and 128 for SDD and SSA.
We use a Mean Squared Error loss for all agents in the scene. To ensure multiple diverse future generation, we use an MSE based Variety Loss~\cite{gupta2018social}, i.e. we optimize only the closest trajectory to the ground truth among the K generated ones. In this way, each read head of the Social Memory Module learns to focus on different locations in memory, thus generating different future modalities.

To avoid overfitting, we perform data augmentation of the training set by applying random rotations to all trajectories in the scene, in the range $[0, 2\pi]$. At the beginning of each episode, memory content is wiped out, both during training and during testing, since we do not want information about an episode to interfere with other samples.

\subsection{Metrics}
\label{sec:metrics}
To evaluate predictions, we adopt the widely used Average Displacement Error (ADE) and Final Displacement Error (FDE) metrics. 
ADE is the average L2 error between all future timesteps, while FDE is the L2 error at the last timestep. In the multimodal configuration, i.e. when we generate multiple diverse predictions, we consider the minimum ADE and FDE between all predictions generated by the model and the ground truth. This is a standard practice, often referred to as best-of-K.

To evaluate the model on SSA, along with ADE/FDE, we observe in which order the agents cross the center of the circumference. The rationale is that social interaction always happen before an agent reaches the opposite quadrant of the circle, since they are all heading towards the center. An incorrect modeling of social interactions will therefore cause the agents not to stop/proceed in the correct order. We used the Kendall's $\tau$ Rank Correlation~\cite{kendall1948rank} to compare the ordering of the agents according to the generated predictions with the ground-truth sequence. 

\revision{Furthermore, we also introduce a new metric to quantitatively assess the capacity of attention-based methods to estimate cause-effect relationships on SSA. The metric, which we call Cause-Effect Accuracy (CEA), evaluates if the model is able to correctly identify which agent causes reactions in the motion patterns of the others. For every moment where an interaction is present (an agent is stopped due to an incoming collision with a faster one), we compare the agent responsible for such interaction with the one estimated by the model and compute the accuracy dividing the number of correct matches with the total number of interactions.}

\minorrevision{
The CEA metric for a an episode can be computed as

\begin{equation}
    CEA = \frac{1}{|A||T|} \sum_{i \neq j, t} \mathbbm{1} \left\{ y^t_{ij} = \hat{y}^t_{ij} \right\} 
\end{equation}

where the true label $y^t_{ij}$  is 1 in case agent $i$ is conditioning the motion of $j$ and 0 otherwise at time $t$, while $\hat{y}^t_{ij}$ is a binary  prediction of such causal relationship between the two agents.

}


\subsection{Results}

We first analyze the capability of SMEMO to address social interactions using SSA. In this synthetic dataset, modeling information such as speed and direction is trivial since agents either move in a straight line or wait for their turn to start moving. The model must learn to reason about reciprocal agent positions to understand when they will interact.
\revision{The distribution of trajectories in SSA and in real datasets like SDD and ETH/UCY will inevitably be different. Trajectories in SSA follow artificial rules that are not meant to reflect real behaviors of pedestrians. On the contrary, the purpose of SSA is to strip down to a bare minimum what agents are allowed to do, removing the need for multimodality and making motion patterns depend solely on social interactions.}

We compare our model against three non-social baselines and five state-of-the-art trajectory predictors. The non-social baselines process each agent individually, without any knowledge about the others. We use a linear regressor, a Multi-Layer-Perception regressor (MLP) and an Encoder-Decoder model based on two Gated Recurrent Units. Note that the Encoder-Decoder model is analogous to the Egocentric Stream in SMEMO.
\revision{We then retrain Expert-Goals~\cite{he2021where}, PECNet~\cite{mangalam2020not}, Trajectron++~\cite{salzmann2020trajectron++}, Social-GAN~\cite{gupta2018social}, AgentFormer~\cite{yuan2021agentformer} and SR-LSTM~\cite{zhang2019sr} on SSA using the official code available online.}
No future multimodality is tested with SSA, since the output is deterministic and we only want to assess the social reasoning capabilities of the model.

In Fig.~\ref{fig:toy_sample} we show a qualitative analysis of an episode from SSA. We compare the ground truth with predictions made by SMEMO and the Encoder-Decoder GRU (i.e. SMEMO's Egocentric Stream). Interestingly, SMEMO does not completely halt trajectories but rather smoothly decreases and increases the velocity of the agents. At the same time, it is able to effectively identify interactions and make slower agents wait for the faster ones to have passed.
The non-social baseline instead is only able to exploit biases in the data distribution to generate average solutions based on the initial speed, e.g. constant velocity trajectories for faster agents and trajectories that slow down when approaching the center for the others.

\begin{figure}[t]
	\centering
	\resizebox{0.5\textwidth}{!}{
		\begin{tabular}{cccccc}
			\multicolumn{6}{c}{\includegraphics[width=.5\textwidth]{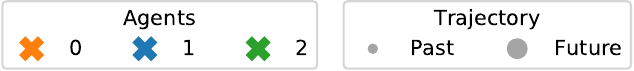}} \\
			
			\multicolumn{2}{c}{\includegraphics[width=.329\textwidth]{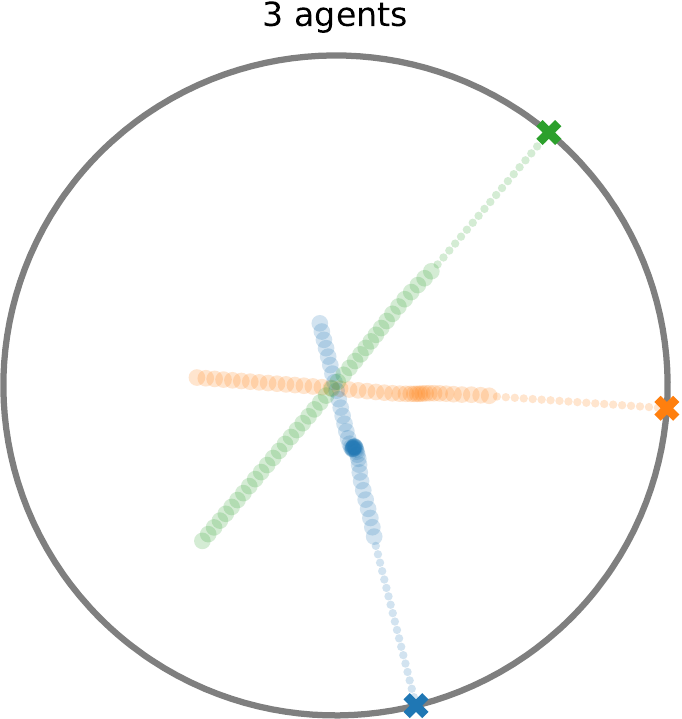}} &
			\multicolumn{2}{c}{\includegraphics[width=.329\textwidth]{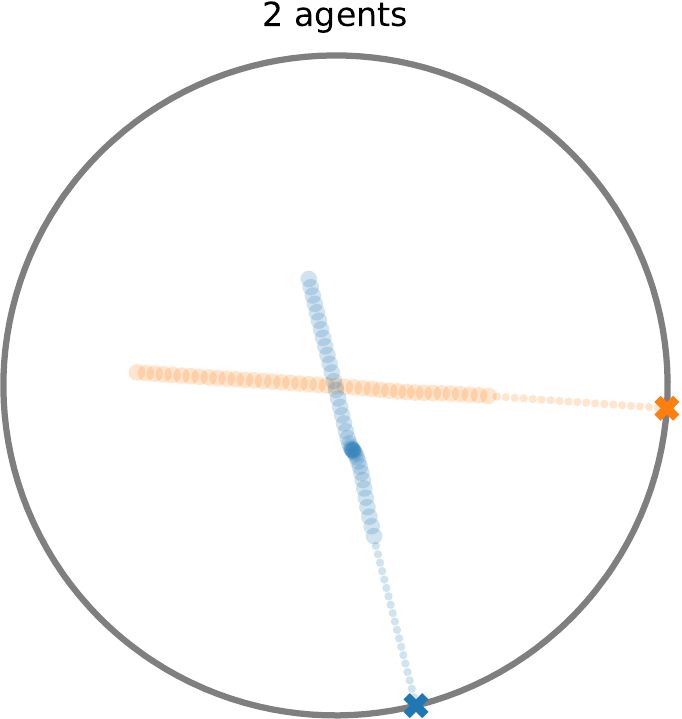}} &
			\multicolumn{2}{c}{\includegraphics[width=.329\textwidth]{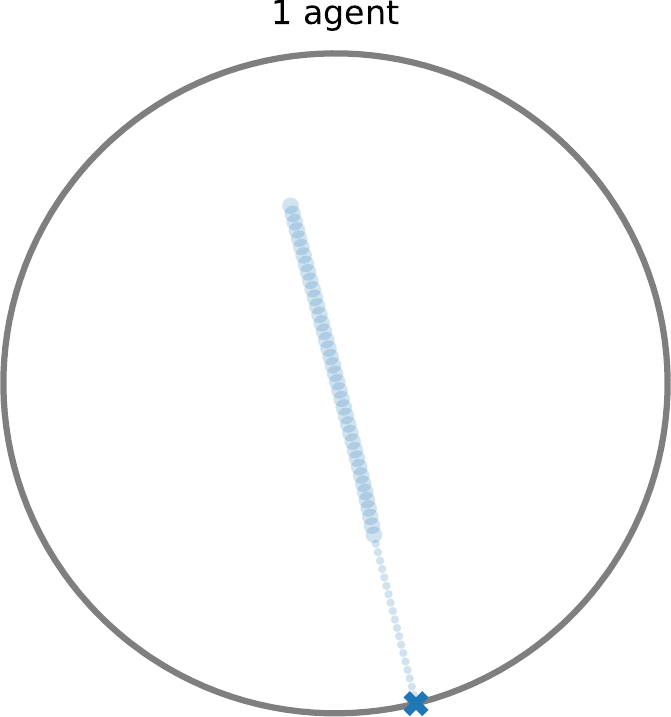}} \\
			
		\end{tabular}	
	}
	\caption{
		\label{fig:removed_agents}Cause-effect modeled by SMEMO. The same episode from SSA is evaluated by iteratively removing the fastest agent. \textit{Left}: the predictions for the blue and orange agents wait for the green one to pass. \textit{Middle}: the orange agent is free to proceed without halting. \textit{Right}: When a single agent is present, SMEMO correctly predicts a constant velocity trajectory.}
\end{figure}

What SMEMO is learning is a chain of cause-effect relationships that makes an agent stop depending on the position of another. This becomes more evident in Fig.~\ref{fig:removed_agents} where we demonstrate how predictions are modified when looking at the same episode with increasingly less agents. The three agents have different speeds, so in the complete episode the fastest does not stop while the other two have to wait. It can be seen that if we remove the fastest agent, the second one keeps its speed until the end while the slowest keeps waiting for its turn. With a single agent instead, no waiting is predicted.

\begin{table}[t]
	\centering
	\caption{\label{tab:toy}\minorrevision{Comparison with the state-of-the art on SSA in terms of ADE, FDE and Kendall's $\tau$ on SSA. SMEMO largely outperforms the competition.}}
	\begin{tabular}{l|c|c|c}
		\multicolumn{1}{c|}{\textbf{Method}} & \multicolumn{1}{c|}{\textbf{ADE}~$\downarrow$} & \multicolumn{1}{c|}{\textbf{FDE}~$\downarrow$} & \textbf{Kendall}~$\uparrow$ \\ \hlineB{3}
		Linear                                & 0.552 ${\scriptstyle \, \pm 0.004}$         & 0.855  ${\scriptstyle \, \pm 0.006}$        & 0.665 ${\scriptstyle \, \pm 0.004}$                        \\
		MLP                                   & 0.527  ${\scriptstyle \, \pm 0.004}$        & 0.832  ${\scriptstyle \, \pm 0.003}$        & 0.638   ${\scriptstyle \, \pm 0.010}$                    \\
		GRU ENC-DEC                           & 0.525 ${\scriptstyle \, \pm 0.004}$          & 0.829 ${\scriptstyle \, \pm 0.003}$         & 0.642  ${\scriptstyle \, \pm 0.009}$                       \\
		Expert-Goals~\cite{he2021where} & 0.571 ${\scriptstyle \, \pm 0.005}$ & 0.896 ${\scriptstyle \, \pm 0.007 }$ & 0.495 ${\scriptstyle \, \pm 0.006}$\\
		PECNet~\cite{mangalam2020not}                                & 0.286  ${\scriptstyle \, \pm 0.012}$        & 0.828  ${\scriptstyle \, \pm 0.009}$       & 0.705       ${\scriptstyle \, \pm 0.003}$                 \\
		Trajectron++~\cite{salzmann2020trajectron++}                          & 0.519   ${\scriptstyle \, \pm 0.011}$        & 0.818  ${\scriptstyle \, \pm 0.019}$         & 0.569    ${\scriptstyle \, \pm 0.015}$                    \\
		Social-GAN~\cite{gupta2018social}                            & 0.302     ${\scriptstyle \, \pm 0.004}$     & 0.506  ${\scriptstyle \, \pm 0.003}$        & 0.626   ${\scriptstyle \, \pm 0.031}$                      \\
		AgentFormer~\cite{yuan2021agentformer} & 0.243 ${\scriptstyle \, \pm 0.003}$ & 0.385 ${\scriptstyle \, \pm 0.003}$ & 0.701 ${\scriptstyle \, \pm 0.006}$
		\\
		SR-LSTM~\cite{zhang2019sr} & 0.217 ${\scriptstyle \, \pm 0.004}$ & 0.409 ${\scriptstyle \, \pm 0.003}$ & 0.777 ${\scriptstyle \, \pm 0.012}$\\
		SMEMO                                 & \textbf{0.169} \textbf{${\scriptstyle \, \pm 0.006}$} & \textbf{0.244} \textbf{${\scriptstyle \, \pm 0.012}$} & \textbf{0.827} \textbf{${\scriptstyle \, \pm 0.008}$}              
	\end{tabular}
\end{table}

\newcolumntype{?}{!{\vrule width 1.5pt}}
\begin{table*}[t]
	\caption{	\label{tab:sdd}\minorrevision{Results on SDD. K is the number of predictions generated by the models. Errors are expressed in pixels. *Results for~\cite{salzmann2020trajectron++} are taken from~\cite{li2020evolvegraph}.}}
	\begin{minipage}[c]{0.4\textwidth}
		\centering
		\begin{tabular}{lcc}
			\multicolumn{3}{c}{K=5}\\ \hline
			\multicolumn{1}{c|}{Method} & \multicolumn{1}{c|}{ADE} & \multicolumn{1}{c}{FDE} \\ \hline
			DESIRE \cite{lee2017desire} & 19.25 & 34.05 \\
			Ridel et al. \cite{ridel2020scene} & 14.92 & 27.97 \\
			MANTRA~\cite{marchetti2020memnet} & 13.51 & 27.34 \\
			PECNet \cite{mangalam2020not} & 12.79 & 25.98 \\
			PCCSNet~\cite{sun2021three} & 12.54 & - \\
			TNT \cite{Zhao2020TNTTT} & 12.23 & 21.16 \\
			\textbf{SMEMO} & \textbf{11.64} & \textbf{21.12} \\
		\end{tabular}
		
	\end{minipage}
	\begin{minipage}[c]{0.59\textwidth}
		\centering
\begin{tabular}{lrr|lrr}
\multicolumn{6}{c}{K=20}\\ \hline
\multicolumn{1}{c|}{Method} & \multicolumn{1}{c|}{ADE} & \multicolumn{1}{c|}{FDE} & \multicolumn{1}{c|}{Method} & \multicolumn{1}{c|}{ADE} & \multicolumn{1}{c}{FDE} \\ \hline
Trajectron++~\cite{salzmann2020trajectron++}* & 19.30                    & 32.70                    & MID~\cite{gu2022stochastic}       & 9.73                    & 15.32                   \\
SoPhie \cite{sadeghian2019sophie}             & 16.27                   & 29.38                   & MANTRA~\cite{marchetti2020memnet} & 8.96                    & 17.76                   \\
EvolveGraph~\cite{li2020evolvegraph}          & 13.90                    & 22.90                    & LB-EBM~\cite{pang2021trajectory}  & 8.87                    & 15.61                   \\
CF-VAE \cite{bhattacharyya2020conditional}    & 12.60                    & 22.30                    & PCCSNet~\cite{sun2021three}       & 8.62                    & 16.16                   \\
P2TIRL \cite{deo2020trajectory}               & 12.58                   & 22.07                   & MemoNet~\cite{xu2022remember}     & 8.56                    & 12.66                   \\
Goal-GAN \cite{Dendorfer_2020_ACCV}           & 12.20                    & 22.10                    & LeapFrog~\cite{mao2023leapfrog}   & 8.48                    & \textbf{11.66}          \\
Expert-Goals~\cite{he2021where}               & 10.49                   & 13.21                   & Y-Net~\cite{mangalam2021goals}    & 8.25                    & 12.10                    \\
SimAug \cite{liang2020simaug}                 & 10.27                   & 19.71                   & \textbf{SMEMO}                    & \textbf{8.11}           & 13.06                   \\
PECNet \cite{mangalam2020not}                 & 9.96                    & 15.88                   &                                   &                         &                        
\end{tabular}


	\end{minipage}
\end{table*}

\newcommand{\red}[1]{{\color{red}{#1}}}
\begin{table}[t]
	\centering
	\caption{
		\label{tab:ethucy}Results on the ETH/UCY datasets. Each model generates K=20 multiple predictions Errors are expressed in meters. Best result for each split in bold; second best is underlined. Methods marked with * exploit map information while other only past trajectories.}
	\resizebox{0.5\textwidth}{!}{
	\begin{tabular}{l|r|r|r|r|r|c}
		\multicolumn{1}{c|}{\textbf{Method}} & \multicolumn{1}{c|}{\textbf{ETH}}        & \multicolumn{1}{c|}{\textbf{HOTEL}} & \multicolumn{1}{c|}{\textbf{UNIV}} & \multicolumn{1}{c|}{\textbf{ZARA1}} & \multicolumn{1}{c|}{\textbf{ZARA2}} & 
		\multicolumn{1}{c}{\textbf{AVERAGE}}\\ \hlineB{3}
		
		SoPhie* \cite{sadeghian2019sophie}                                 & 0.70/1.43                      & 0.76/1.67                 & 0.54/1.24                & 0.30/0.63                 & 0.38/0.78 & 0.54/1.15                 \\
		Next*~\cite{liang2019peeking} & 0.73/1.65 & 0.30/0.59  & 0.60/1.27 & 0.38/0.81 & 0.31/0.68 & 0.46/1.00 \\
		S-BiGAT*~\cite{Kosaraju2019BIGAT}                                & 0.69/1.29                      & 0.49/1.01                 & 0.55/1.32                & 0.30/0.62                  & 0.36/0.75 &  0.48/1.00           \\
		GOAL-GAN*~\cite{Dendorfer_2020_ACCV}                               & 0.59/1.18                      & 0.19/0.35                 & 0.60/1.19                & 0.43/0.87                 & 0.32/0.65  & 0.43/0.85                \\
		Introvert*~\cite{shafiee2021introvert} & 0.42/0.70 & 0.11/0.17 & 0.20/0.32 & 0.16/0.27 &	0.16/0.25 & 0.21/0.34\\ \hline \hline
		
		Social-GAN \cite{gupta2018social}                                  & 0.81/1.52                      & 0.72/1.61                 & 0.60/1.26                & 0.34/0.69                 & 0.42/0.84 & 0.58/1.18                 \\
		CGNS \cite{li2019cgns}                                   & 0.62/1.40                       & 0.70/0.93                  & 0.48/1.22                & 0.32/0.59                 & 0.35/0.71   & 0.49/0.97               \\
		SR-LSTM~\cite{zhang2019sr} &		0.63/1.25 & 0.37/0.74 & 0.51/1.10 & 0.41/0.90 & 0.32/0.70 & 0.45/0.94 \\
		MATF \cite{zhao2019multi}                                   & 1.01/1.75                      & 0.43/0.80                  & 0.44/0.91                & 0.26/0.45                 & 0.26/0.57      & 0.48/0.90              \\
		
		STGAT~\cite{Huang2019stgat} &0.65/1.12&	0.35/0.66	&0.52/1.10&	 0.34/0.69&	0.29/0.60&	0.43/0.83\\
		
		SGCN~\cite{shi2021sgcn} &0.63/1.03&	0.32/0.55	&0.37/0.70&	 0.29/0.53&	0.25/0.45&	0.37/0.65\\
		MANTRA~\cite{marchetti2020memnet} & 0.48/0.88 & 0.17/0.33 & 0.37/0.81 & 0.27/0.58 & 0.30/0.67 & 0.32/0.65 \\		
		Transformer \cite{giuliari2020transformer}                            & 0.61/1.12                      & 0.18/0.30                 & 0.35/0.65                & 0.22/0.38                 & 0.17/0.32 & 0.31/0.55             \\
		PECNet \cite{mangalam2020not}                                 & 0.54/0.87 & 0.18/0.24                 & 0.35/0.60                & 0.22/0.39                 & 0.17/0.30  & 0.29/0.48                \\
		PCCSNet~\cite{sun2021three} & \textbf{0.28}/{0.54} & {0.11}/{0.19} & 0.29/0.60 & 0.21/0.44 &	0.15/0.34 &	{0.21}/0.42\\
		Trajectron++ \cite{salzmann2020trajectron++}                           & 0.39/{0.83}                      & {0.12}/{0.19}                & {0.22}/{0.43}                & {0.17}/0.32                & \textbf{0.12}/{0.25} & {0.20}/{0.40}                \\
		Social-NCE~\cite{liu2021social} & ~~-~~/0.79 & ~~-~~/{0.18} & ~~-~~/0.44 & ~~-~~/0.33 & ~~-~~/0.26 & ~~~-~~/0.40\\
		AgentFormer~\cite{yuan2021agentformer} & 0.45/0.75 & 0.14/0.22 & 0.25/0.45 & {0.18}/\textbf{0.30} &	{0.14}/\textbf{0.24} &	0.23/0.39\\
		LB-EBM~\cite{pang2021trajectory} & {0.30}/\textbf{0.52}&	{0.13}/0.20&	0.27/0.52&	0.20/0.37	&0.15/0.29&	{0.21}/0.38\\
		Expert-Goals~\cite{he2021where} & {0.37}/0.65 & \textbf{0.11}/\textbf{0.15} &	\textbf{0.20}/0.44 &	\textbf{0.15}/{0.31} &	\textbf{0.12}/0.25 &	\textbf{0.19}/0.36\\
		\textbf{SMEMO}                                  & {0.39}/{0.59}                      & {0.14}/{0.20}                 & {0.23}/\textbf{0.41}                & {0.19}/{0.32}                 & {0.15}/{0.26}  &  {0.22}/\textbf{0.35}               \\ 
		
	\end{tabular}
}
\end{table}

In Tab.~\ref{tab:toy} we show numerical results on SSA. \minorrevision{We report mean and standard deviation for all models, which have been trained three times with different random seeds.} As expected, the non-social baselines fail to produce satisfactory predictions. Surprisingly, instead, state-of-the-art methods such as PECNet~\cite{mangalam2020not} and Trajectron++~\cite{salzmann2020trajectron++} achieve similar performances in terms of FDE and Expert-Goals~\cite{he2021where}, which achieves the best results one real data as further discussed in this Section, performs even worse than non-social baselines in all metrics.
We believe that the issue with methods such as~\cite{mangalam2020not} and~\cite{he2021where} lies in the fact that they heavily rely on estimating an endpoint spatial goal. This is indeed an important aspect when modeling real-world trajectories, but in SSA it does not offer any kind of advantage since the destination is always a point along a straight line. What emerges from this analysis is that, while these methods are extremely powerful predictors (see Tab.~\ref{tab:sdd} and Tab.~\ref{tab:ethucy} and discussion further on in this section), they are not equipped with a sufficiently effective social reasoning mechanism.
Social-GAN~\cite{gupta2018social}, on the other hand, is capable of obtaining a lower FDE, yet the Kendall $\tau$ is still on par with non-social methods.
\minorrevision{AgentFormer~\cite{yuan2021agentformer} and SR-LSTM~\cite{zhang2019sr} perform better than the other state-of-the-art methods in ADE and FDE but the results are worse than our model. These two models, similarly to ours, perform an inter-agent attention for every future timestep. We believe that this enables a certain degree of social reasoning while generating the predictions.
However, SMEMO achieves much lower ADE and FDE as well as a considerable improvement in rank correlation, getting close to the upper bound of 1.}

\newcommand{\qualw}{.22\textwidth}
\begin{figure*}[t]
	\centering
	\begin{tabular}{ccc|c}
		
		\includegraphics[width=\qualw, trim=0 0 0 130, clip]{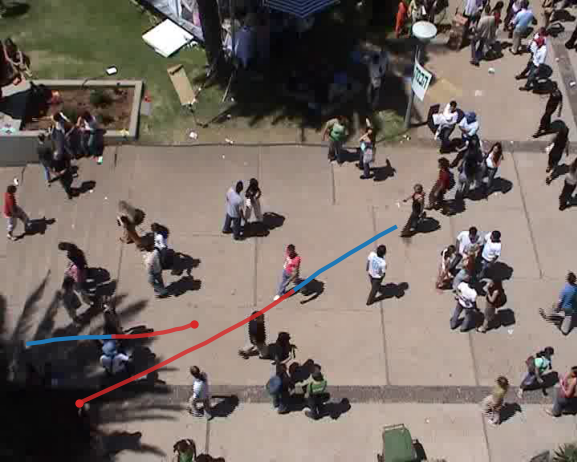} &
		\includegraphics[width=\qualw, trim=0 0 0 130, clip]{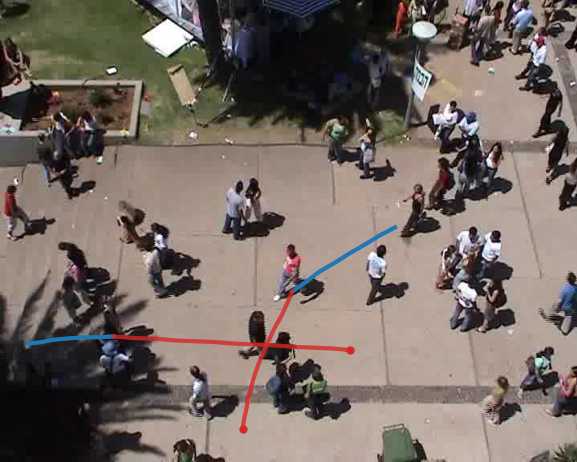} &
		\includegraphics[width=\qualw, trim=0 0 0 130, clip]{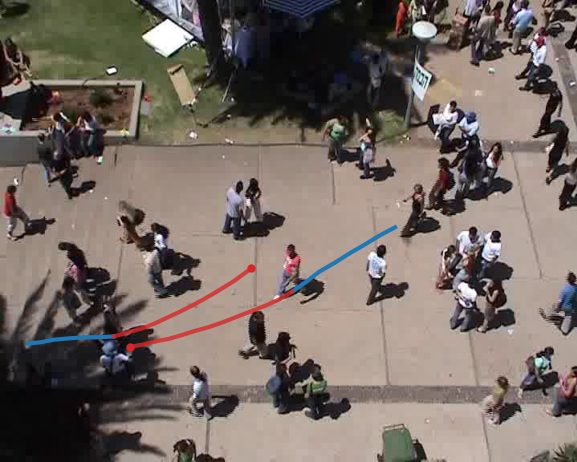} &
		\includegraphics[width=\qualw, trim=0 0 0 130, clip]{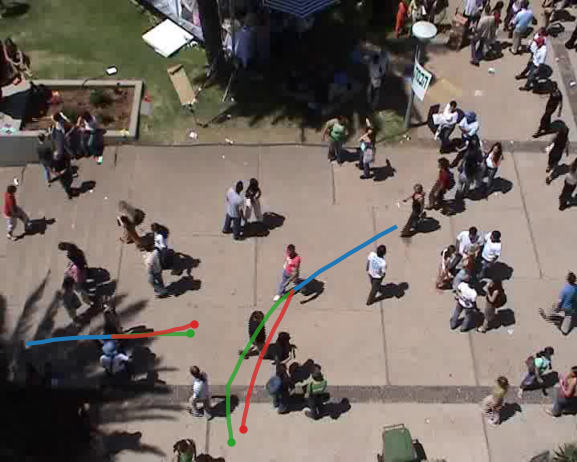}\\
		
		\includegraphics[width=\qualw,trim=0 0 0 130,  clip]{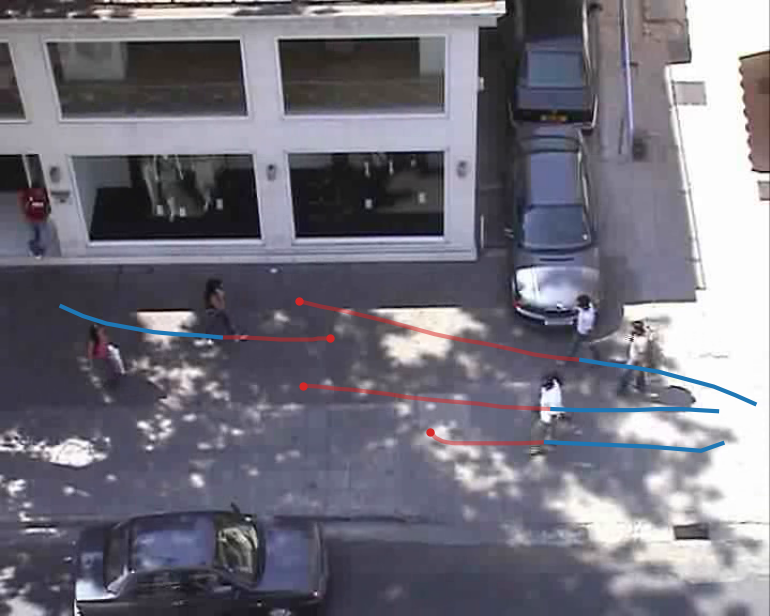} &
		\includegraphics[width=\qualw,trim=0 0 0 130,  clip]{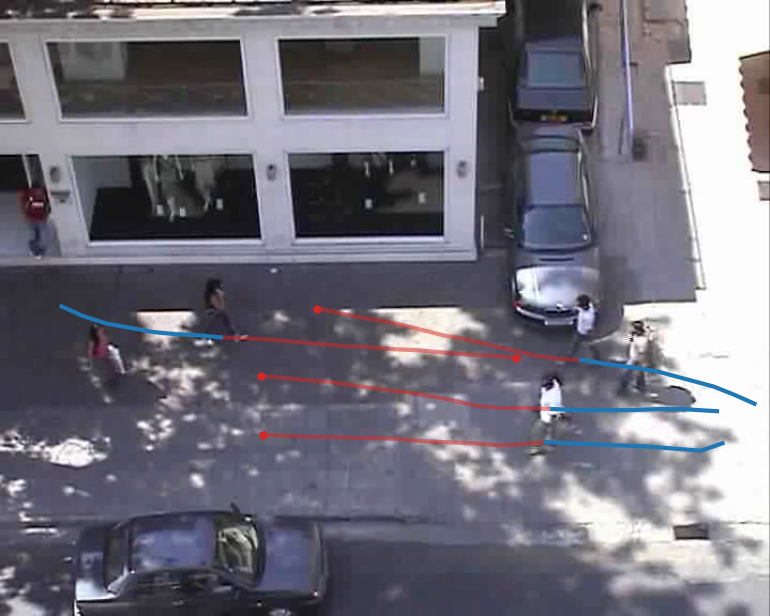} &
		\includegraphics[width=\qualw,trim=0 0 0 130,  clip]{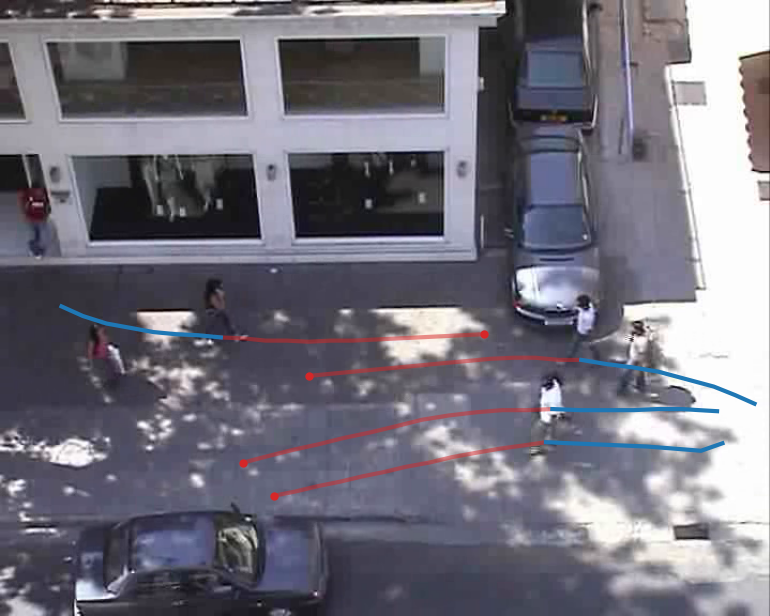} &
		\includegraphics[width=\qualw,trim=0 0 0 130, clip]{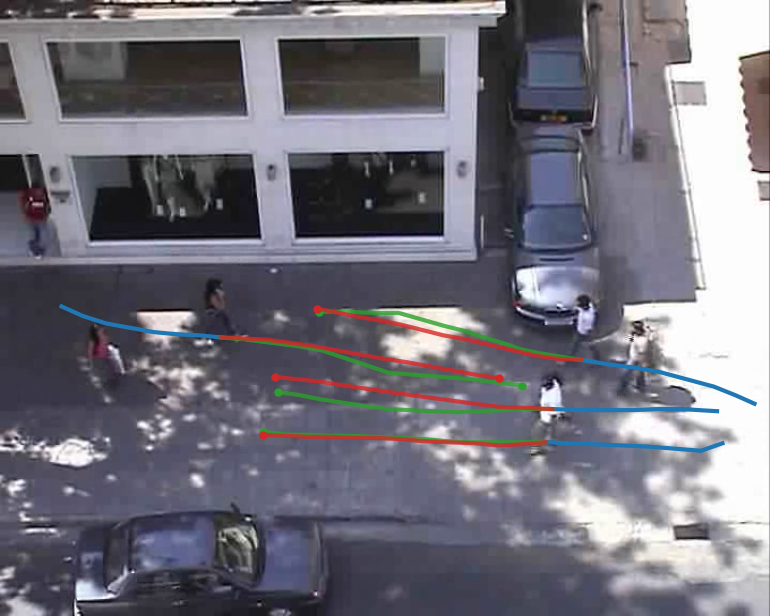}\\
		
		\includegraphics[width=\qualw, trim=0 0 0 50,clip]{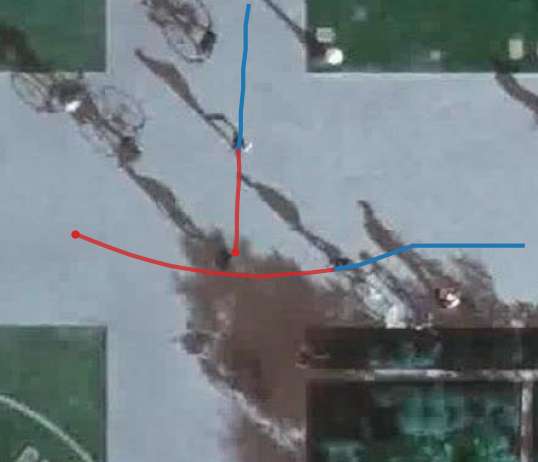} &
		\includegraphics[width=\qualw, trim=0 0 0 50,clip]{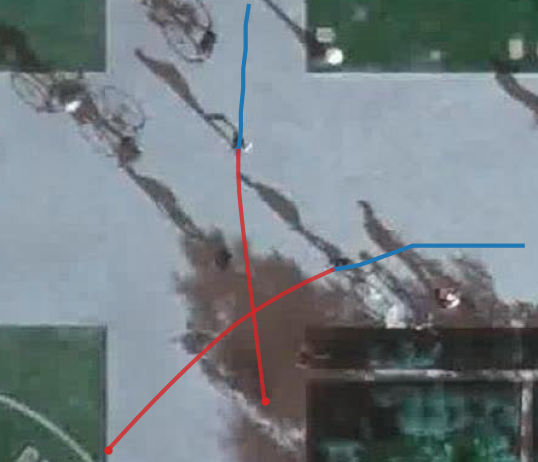} &
		\includegraphics[width=\qualw,trim=0 0 0 50, clip]{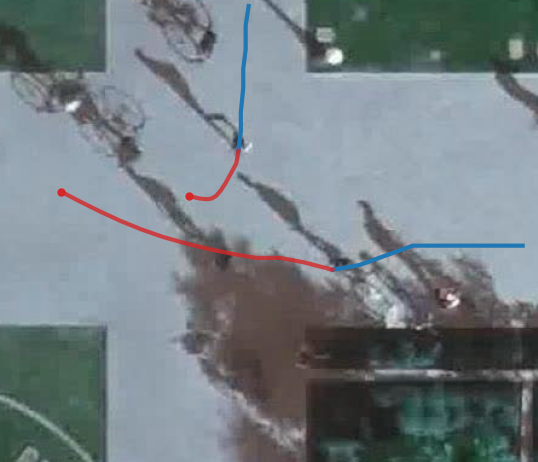} &
		\includegraphics[width=\qualw,trim=0 0 0 50, clip]{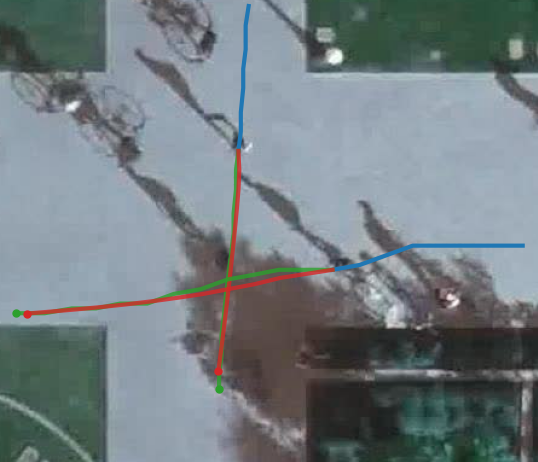}\\

		& (a) Social Multimodal Prediction  & &(d) GT + best prediction\\
		
	\end{tabular}	
	\caption{Social multimodal prediction generated by SMEMO in different real dataset.
		Past trajectory is depicted in blue, ground truth in green and prediction in red. The first and second row show example from ETH/UCY datasets (respectively Univ and Zara scenario), while the third from SDD.
		The first three columns represent different combination of multi-modal prediction generated by model (each combination is related with a specific reading controller), while the last column shows the best prediction compared with ground-truth.
		\label{fig:qualitative}}
\end{figure*}

Moving to real world data, we report the results obtained by SMEMO on SDD and ETH/UCY in Tab.~\ref{tab:sdd} and Tab.~\ref{tab:ethucy}, along with the best performing methods from the state of the art\footnote{Results for \cite{he2021where} do not correspond with the original paper due to an error pointed out by the authors here: \url{https://github.com/JoeHEZHAO/expert_traj}}.
On both datasets we show ADE and FDE for K=20 predictions. Since some prior work also adopts K=5 on SDD, we evaluate our model also with this reduced number of predictions.

\minorrevision{On the SDD dataset, SMEMO obtains state-of-the-art results, except for FDE at $K=20$, where it reports competitive results with the top three performing methods.}

On the ETH/UCY dataset, instead, SMEMO is able to perform better than most of the existing models and comparably with latest approaches. In Tab. \ref{tab:ethucy} we also show results obtained by methods that rely on map information. We do not directly compare to them since, as most methods, we do not rely on such input, yet it is interesting to note that SMEMO is still able to obtain better or on par results (Introvert~\cite{shafiee2021introvert}).
Even if there is not a predominance of a single state-of-the-art model for each split of the dataset, SMEMO manages to achieve excellent results on the average of all splits having the best result in the FDE.

On both datasets, the only real competitor appears to be Expert-Goals~\cite{he2021where}, which achieves similar results to SMEMO. In particular, while SMEMO is able to lower the FDE considerably on SDD dataset and to equal it on ETH/UCY, Expert-Goals~\cite{he2021where} exhibits a better ADE, i.e. a better short term behavior.
We argue however that the most challenging aspects of trajectory prediction appear when generating long term predictions, which are the most likely to be influenced by social interactions.

In Fig. \ref{fig:qualitative}, we report some real qualitative examples from the ETH/UCY and SDD dataset. As we can see from the different combinations generated by the model, SMEMO is able to generate multiple predictions that are consistent with both the agent's past trajectory and its social context.

\begin{table}[]
	\centering
		\caption{\label{tab:toy_ablation}\minorrevision{Ablation study on SSA. \textit{Memory reset}: content wiped out at present. \textit{Zero/Random reading}: read heads always read zero/random vectors. \textit{State pooling}: no external memory; pooled controller states for each agent are fed to the egocentric stream.}}
	\begin{tabular}{l|c|c|c}
		\textbf{Ablation study} & \multicolumn{1}{l|}{\textbf{ADE}~$\downarrow$} & \multicolumn{1}{l|}{\textbf{FDE}~$\downarrow$} & \textbf{Kendall}~$\uparrow$ \\ \hlineB{3}
		SMEMO          & \textbf{0.169}                    & \textbf{0.244}                    & \textbf{0.827}        \\
		Memory reset            & 0.180                             & 0.270                             & 0.790                 \\
    State pooling          & 0.270                   & 0.414                         & 0.690 \\
		Zero reading           & 0.522                             & 0.822                             & 0.640                 \\
		Random reading          & 0.522                             & 0.822                             & 0.650                
	\end{tabular}
\end{table}

\begin{table*}[]
	\centering
	\caption{\label{tab:real_ablation}Ablation study on SDD and ETH/UCY. \textit{Memory reset}: content wiped out at present. \textit{Zero/Random reading}: read heads always read zero/random vectors. \textit{State pooling}: no external memory; pooled controller states for each agent are fed to the egocentric stream. We report both ADE and FDE.}
	\begin{tabular}{l|c|c||c|c|c|c|c|c}
		\textbf{Ablation study} & \multicolumn{1}{l|}{\textbf{SDD (K=5)}} & \multicolumn{1}{l||}{\textbf{SDD(K=20)}} & \textbf{ETH} & \textbf{HOTEL} & \textbf{UNIV} & \textbf{ZARA1} & \textbf{ZARA2} & \textbf{AVERAGE} \\ \hlineB{3}
		SMEMO          & \textbf{11.64/21.12}                    & \textbf{~~8.11/13.06}                    & \textbf{0.39/0.59}   & \textbf{0.14/0.20} & \textbf{0.23/0.41} & \textbf{0.19/0.32} & \textbf{0.15/0.26} & \textbf{0.22/0.35}    \\

		Memory reset            & 12.29/22.51                          & ~~8.43/13.43                           & 0.46/0.72    & 0.16/0.28 & 0.25/0.45 & 0.20/0.35 & 0.17/0.29 & 0.24/0.42            \\
		Zero reading           & 19.19/35.66                            & 19.30/35.87                             & 0.93/1.91   & 0.27/0.52 & 0.51/1.11 & 0.42/0.95 & 0.32/0.76 & 0.49/1.05              \\
		Random reading          & 14.89/28.54                             & ~~9.48/15.39                             & 0.53/0.95  & 0.18/0.27 & 0.29/0.55 & 0.26/0.50 & 0.20/0.40 & 0.29/0.53              \\
		State pooling          & 17.77/35.39                  & 17.93/35.75                     & 0.93/1.90 & 0.28/0.54 & 0.51/1.11 & 0.43/0.95 & 0.32/0.71 & 0.49/1.04
	\end{tabular}
\end{table*}

\subsection{Ablation Studies}
\revision{\textbf{Importance of Memory} We perform ablations studies to assess the importance of the external working memory by retraining SMEMO with the following variations: i) \textit{Memory Reset} - the memory gets wiped out as soon as the present timestep is reached; ii) \textit{Zero reading} - the memory reading controller always reads a vector of zeros; iii) \textit{Random reading} - the memory reading controller always reads a random vector; iv) \textit{State Pooling} - the Social Memory Module just outputs the state of the memory controller, average-pooled among all agents.}
\revision{Tab. \ref{tab:toy_ablation} and Tab.~\ref{tab:real_ablation} we report the results in terms of ADE, FDE respectively for the synthetic and real datasets. For SSA we also report Kendall $\tau$ Rank Correlation.
Interestingly, resetting the memory at the present timestep does not affect too much the final results. All the memory about the past is condensed into the latent state of the recurrent encoder and controller. In particular on SSA, since no social interaction takes place in the past, the model is still able to re-write relevant information in memory and perform quite well.
Zero and random reading obtain almost identical results in Tab.~\ref{tab:toy_ablation} since the decoder learns to ignore the input generated by the Social Memory Module, which does not carry any relevant information. As expected, this performs on par with the GRU Encoder-Decoder baseline from Tab. \ref{tab:toy}.}

\revision{The reason of this behavior is also to be found in the fact that on SSA we predict a single future. In fact, moving to the real datasets, where we perform multimodal predictions, reading random states allows us to generate diverse futures, thus lowering the prediction error. It can be seen in Tab.~\ref{tab:real_ablation} that for SDD the gap between zero reading and random reading increases when predicting more futures.
In simple terms, the zero reading setting makes all the predictions collapse into the same output since all the memory read heads extract the same information.
A similar behavior can be observed for state pooling. When the memory controller outputs an average-pooled state among agents, multimodality is lost.
However, in this case we are relying on an internal memory model instead of an external one. On SSA this leads to double ADE and FDE with Kendall $\tau$ dropping to similar values to the non-social baselines. This demonstrates how the model can benefit from relying on a working memory in which data can be explicitly stored instead of blended in a unique latent vector.}

\begin{table*}[t]
	\centering
	\caption{Results on the SDD and ETH/UCY datasets using a social-level GRU to update the memory content instead of SMEMO's update policy (Eq.~\ref{eq:update}). \label{tab:gru_update}}
	\begin{tabular}{l|c|c||c|c|c|c|c|c}
		\textbf{Memory Update} & \multicolumn{1}{c|}{\textbf{SDD K=5}} & \multicolumn{1}{c||}{\textbf{SDD K=20}} & \textbf{ETH} & \textbf{HOTEL} & \textbf{UNIV} & \textbf{ZARA1} & \textbf{ZARA2} & \textbf{AVERAGE}\\ \hlineB{3}
		SMEMO & \textbf{11.64/21.12} & \textbf{8.11/13.06}  & \textbf{0.39/0.59} 
		& \textbf{0.14/0.20}  & \textbf{0.23/0.41}  & \textbf{0.19/0.32}  & \textbf{0.15/0.26}  & \textbf{0.22/0.35}\\
		GRU  & 11.80/21.25 & 8.27/13.69  & 0.43/0.72 & 0.16/0.26 & 0.25/0.46 & 0.19/0.35 & 0.16/0.29 & 0.24/0.42
	\end{tabular}
\end{table*}

\revision{\textbf{Memory Update}
Eq.~\ref{eq:update} states that memory content is updated sequentially, generating the new memory $\textbf{M}_{t+1}$ depending on the previous content $\textbf{M}_t$ and new information to be added ($\textbf{A}_t$) or deleted ($\textbf{E}_t$).
Theoretically, this mechanism is similar to that used by a recurrent neural network such as an LSTM or GRU: the write matrix $\textbf{A}_t$ resembles the information passing through an input gate, while the effect of the erase matrix $\textbf{E}_t$ is similar in spirit to a forget gate.
A recurrent neural network could indeed be defined using similar equations. Therefore, we study the effects of using a GRU to update the social memory content.}
\revision{To do so, we treat the content of the external memory as the internal state of the GRU, unrolling its cells into a single mono-dimensional vector. This vector will have size $|\textbf{M}|*Q$ where $|\textbf{M}|$ is the number of cells and $Q$ is the dimension of the features stored in memory.
At each timestep, a vector of length $|\textbf{M}|*Q$ is fed to the GRU. We generate it in the same way as the add and erase vectors. For simplicity we generate a single \textit{update} rather than two distinct \textit{add} and \textit{erase} vectors, as the GRU requires a single input.
After the update, the internal state of the GRU is reshaped back to the original external memory size $|\textbf{M}| \times Q$, i.e., with $|\textbf{M}|$ cells of dimension $Q$. We leave the reading phase unaltered.}
\revision{We test the social-level GRU to update the memory on the SDD and ETH/UCY datasets and report the results in Tab.~\ref{tab:gru_update}. Overall, the results are close to the ones obtained with the original memory update formulation, although worse results are obtained with the GRU. We impute this drop to the lack of individual addressing of memory cells during the update process.}

\revision{\textbf{Integration with Other Models}
In principle, the Social Memory Module could be plugged in into any existing trajectory prediction model. In particular, if a model has an encoder-decoder structure to process single trajectories it could benefit from SMEMO to analyze the social context since encoded trajectories could directly feed SMEMO's controllers. In the same way, the social feature produced by SMEMO can be used to condition future generation by feeding it to the decoder.}

\revision{To demonstrate the effectiveness and integrability of our approach, we augmented a model from the state of the art, PECNet~\cite{mangalam2020not}, with a Social Memory Module.
PECNet originally exploited a social pooling mechanism, which we replaced in favor of SMEMO. We also updated PECNet's decoder to make the prediction generation autoregressive, which is necessary for the Social Memory Module since data is continuously read and written in memory at every timestep.
We carried out the experiment on the SDD dataset with K=20 futures, by retraining the augmented model. We report results in Tab.~\ref{tab:smemo_pecnet}.
It can be seen that both for ADE and FDE there is approximately a $4.5\%$ improvement against the original Social Pooling mechanism.}

\begin{table}[t]
	\centering
	\caption{Results on the SDD dataset for K=20 futures obtained by the PECNet model~\cite{mangalam2020not} using a Social Memory Module (SMEMO) instead of its original Social Pooling Mechanism (SP). We also report a baseline without social information to understand the effect of the two modules. \label{tab:smemo_pecnet}}
	\begin{tabular}{c|c|c|c}
		\textbf{K=20} & \textbf{PECNet w/o Social} & \textbf{PECNet + SP} & \textbf{PECNet + SMEMO} \\ \hlineB{3}
		FDE & 16.72 & 15.88 & \textbf{15.15} \\ 
		ADE & 10.56  & 9.96  & \textbf{9.38} 
	\end{tabular}
\end{table}

\begin{table}[t]
	\centering
	\caption{Comparison of SMEMO with and without context (top-view semantic map of the surronding scene). \label{tab:with_context}}
	\begin{tabular}{c|c|c}
		\textbf{SMEMO (ADE/FDE~$\downarrow$)} & \multicolumn{1}{l|}{\textbf{SDD (K=5)}} & \multicolumn{1}{l|}{\textbf{SDD (K=20)}}  \\ \hlineB{3}
		w/o context          & 11.64/21.12                  & \textbf{8.11}/13.06\\
		with context            & \textbf{11.41/20.66}                          & 8.24/\textbf{13.03}
	\end{tabular}
\end{table}

\begin{table}[t]
	\centering
	\caption{Execution time and space occupancy of SMEMO and its components.\label{tab:time}}
	\begin{tabular}{lcc}
		& K=5 & K=20 \\ \hlineB{3}
		\multicolumn{3}{c}{Time (ms)} \\ \hline
		\multicolumn{1}{l|}{Encoding} & \multicolumn{1}{c|}{12 (12.7\%) } & 22 (14.2\%) \\
		\multicolumn{1}{l|}{Reading} & \multicolumn{1}{c|}{28 (29.7\%)} & 66 (42.8\%) \\
		\multicolumn{1}{l|}{Writing} & \multicolumn{1}{c|}{51 (54.2\%)} & 53 (34.4\%) \\
		\multicolumn{1}{l|}{Decoding} & \multicolumn{1}{c|}{7 (7.4\%)} & 13 (8.4\%) \\
		\multicolumn{1}{l|}{Total} & \multicolumn{1}{c|}{94 (100\%)} & 154 (100\%) \\ \hline \hline
		\multicolumn{3}{c}{Size (Mb - \%)} \\ \hline
		\multicolumn{1}{l|}{Encoder-Decoder} & \multicolumn{1}{c|}{0.629 (89.4\%) } & 0.629 (76.3\%) \\
		\multicolumn{1}{l|}{Social Memory Module} & \multicolumn{1}{c|}{0.064 (9.1\%)} & 0.185 (22.4\%) \\
		\multicolumn{1}{l|}{External Memory} & \multicolumn{1}{c|}{0.010 (1.4\%)} & 0.010 (1.2\%) \\
		\multicolumn{1}{l|}{Total} & \multicolumn{1}{c|}{0.703 (100\%)} & 0.824 - 100\%)
	\end{tabular}
\end{table}

\revision{\textbf{Including the Environmental Context}
The focus of this work is on analyzing the social aspect of trajectory prediction, therefore we disregarded the usage of the environmental context, i.e. a map of the surrounding scene. However, this could be easily plugged in into the model.
We performed an experiment adding a CNN processing the semantic top-view map of the scene in which the agents move. The map encodes information about roads, sidewalks, vegetation and buildings. For each agent, in the timestep corresponding to the present, we take a crop of 200x200px of the context centered on its position.
The CNN is composed of three convolutional layers. Each layer has a ReLU activation and a max-pooling. All layers have a 3x3 kernel and padding 1. The first two convolutional layers have stride 1 and the third one stride 2. In the end, a linear function generates a feature vector that represents the content of the map.

The obtained feature is concatenated to the feature generated by the agent's movement and by the one read from the memory, for each timestep. All three are thus fed to the decoder which makes predictions conditioned by the surrounding map as well as the social context.
We performed the experiments on the SDD dataset for K=5 and K=20, obtaining a slight improvement in the results, especially for K=5. Results are reported in Tab. \ref{tab:with_context}.}

\revision{\textbf{Execution Time Analysis}
We have measured the computational time during the inference of SMEMO. The number of predicted futures has an impact so we carried out the analysis for K=5 and K=20 using an Nvidia GeForce RTX2080.
As can be seen in Tab.~\ref{tab:time}, on average the most computationally expensive phases are memory reading and writing (42.8\% and 34.41\% in case of K=20). Whereas writing does not depend on the number of futures, execution time for reading depends on the number of heads used for multimodal predictions.
In addition, we report the size of the SMEMO model in Megabytes.
The Social Memory Module does not significantly impact the overall size of the model (22\% for K=20, 9\% for K=5).
Furthermore, the external memory occupies only 0.01Mb, i.e. the 1.2\% of the entire model, when using a memory size $|\textbf{M}|=128$.}

\begin{figure}[t]
	\centering
	\setlength\tabcolsep{1pt} 
	\begin{tabular}{cc}
		\multicolumn{2}{c}{\includegraphics[width=.8\columnwidth]{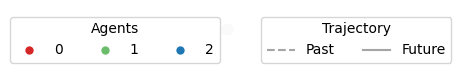}} \\
		\begin{tabular}{c}
			\includegraphics[width=0.35\columnwidth]{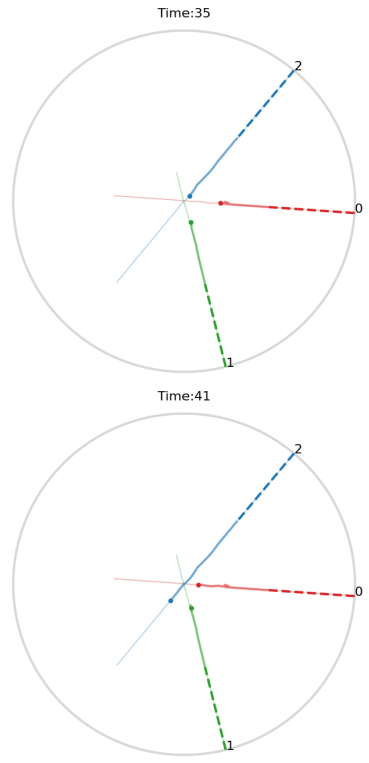}
		\end{tabular}
		&	
		\begin{tabular}{cc}
			\includegraphics[width=0.5\columnwidth]{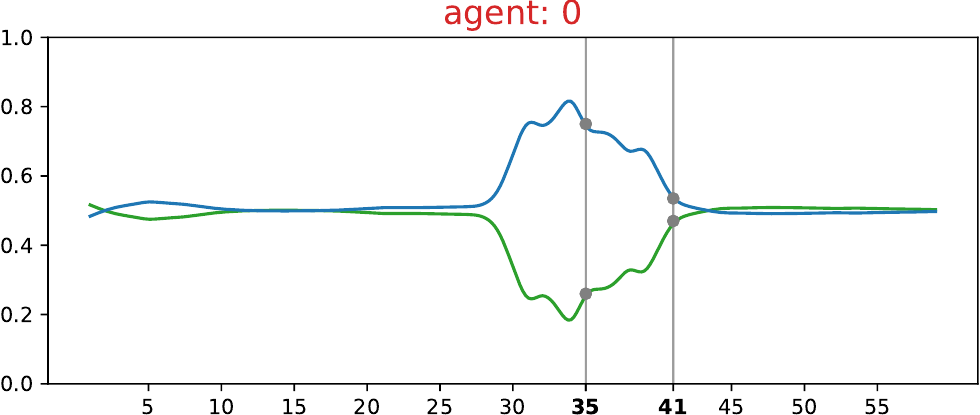} \\ 
			\includegraphics[width=0.5\columnwidth]{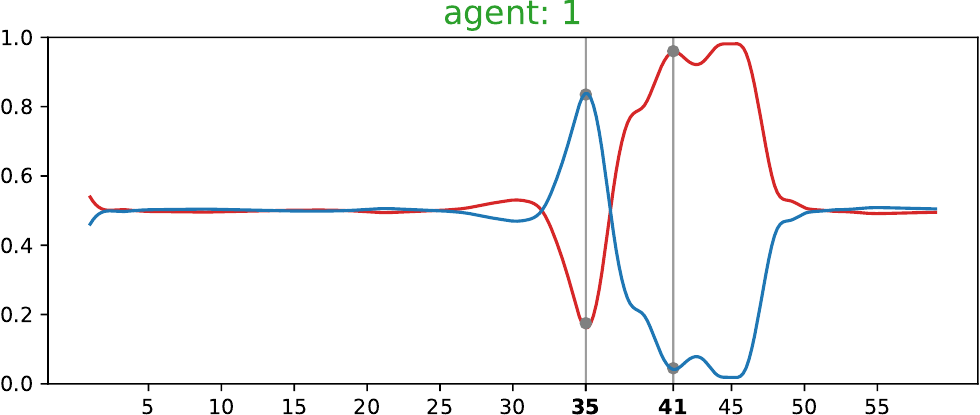} \\
			\includegraphics[width=0.5\columnwidth]{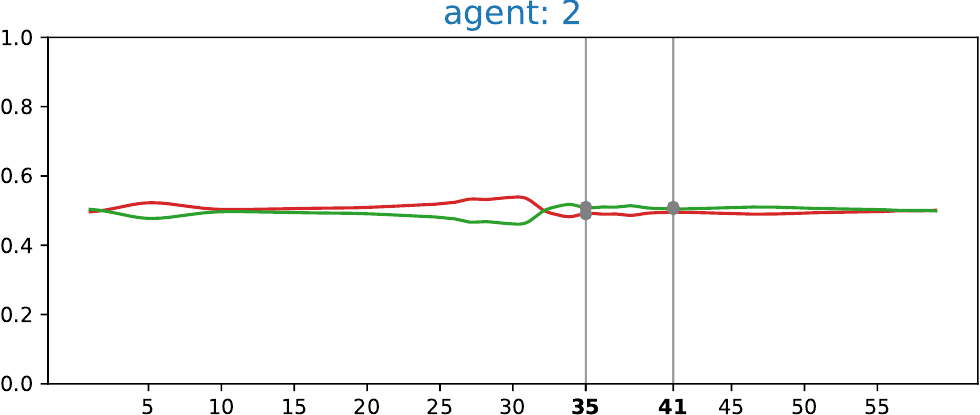} \\ 
		\end{tabular}
		\\
		
	\end{tabular}
	\caption{
		\label{fig:exp_toy}Explainability analysis of an example from SSA. \textit{Left}: agent trajectories at t=35 and t=41. Past and future are shown with a dashed and thick line respectively. The thin line represents the complete trajectory up to the final prediction horizon (t=60). \textit{Right}: SMEMO's reading attention for each agent on the others for each timestep. }
	
\end{figure}

\subsection{Explainability Results}
\label{sec:explainability_res}
\begin{figure}[t]
	\centering
	\setlength\tabcolsep{1pt} 
	\begin{tabular}{cc}
		\multicolumn{2}{c}{\includegraphics[width=.8\columnwidth]{img/exp/legend}} \\
		\begin{tabular}{c}
			\includegraphics[width=0.45\columnwidth]{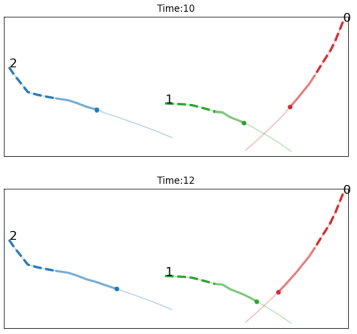}
		\end{tabular}
		&
		\begin{tabular}{cc}
			\includegraphics[width=0.5\columnwidth]{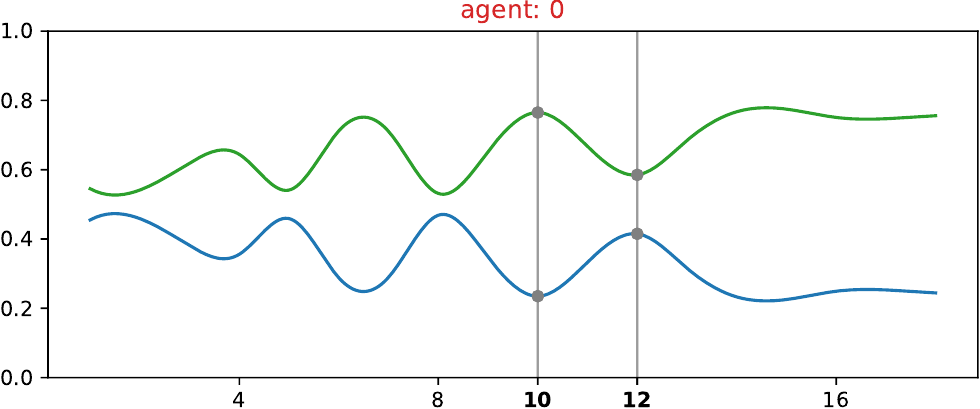} \\ 
			\includegraphics[width=0.5\columnwidth]{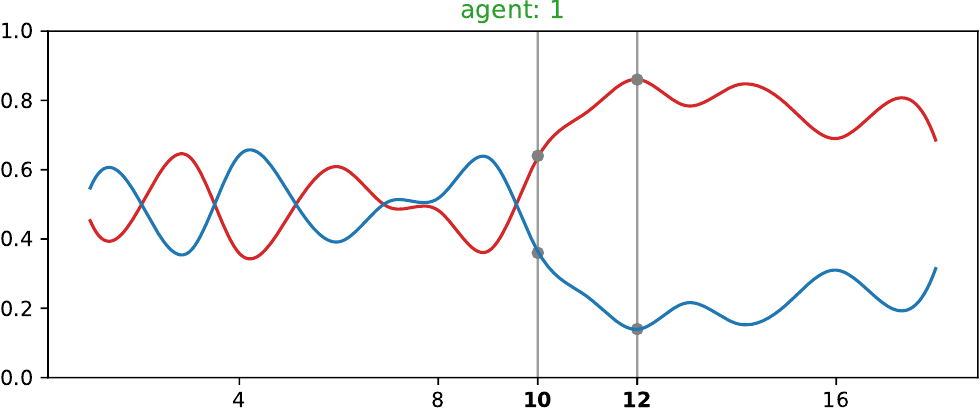} \\
			\includegraphics[width=0.5\columnwidth]{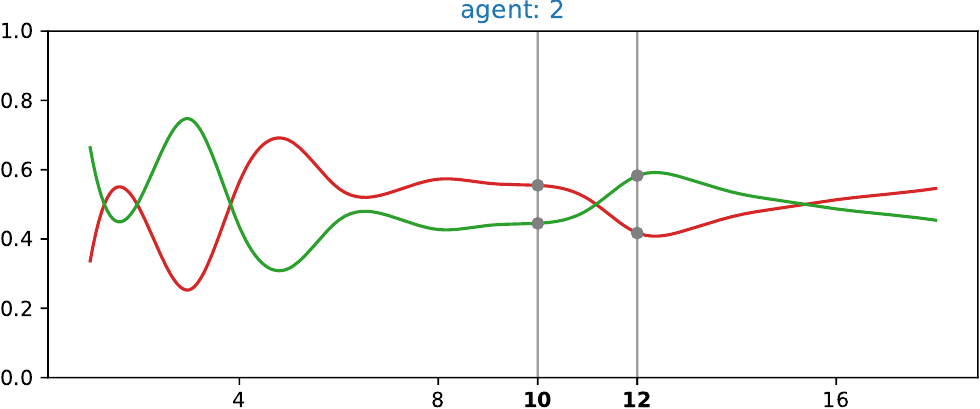} \\ 
		\end{tabular}
		\\
		
	\end{tabular}
	\caption{
		\label{fig:exp_real}Explainability analysis of an example from ETH/UCY. \textit{Left}: agent trajectories at t=10 and t=12. Past and future are shown with a dashed and thick line respectively. The thin line represents the complete trajectory up to the final prediction horizon (t=20). \textit{Right}: SMEMO's reading attention for each agent on the others for each timestep. }
	
\end{figure}

To demonstrate SMEMO's capabilities to provide explainable predictions, as described in Sec.~\ref{sec:exp}, we report two examples, a SSA scenario (Fig. \ref{fig:exp_toy}) and a real-world scenario taken from ETH/UCY (Fig. \ref{fig:exp_real}). In both examples three agents are present and their trajectories are displayed focusing on two different timesteps that exhibit interesting social interactions.

Along with the predicted trajectories, we show SMEMO's reading attentions during the whole episode. For each agent $i$ we plot $\textbf{att}^i(j)$ for each $j \neq i$, highlighting how inter-agent attentions change through time.
Since attention values are normalized with a softmax, all attentions sum up to 1 at each timestep. As a consequence, when no relevant social interaction is present, attentions for agent $i$ over the others will have similar values that oscillate around $\textbf{att}^i(j)=1/(N-1)$, i.e. one divided by the number of agents in the scene excluded $i$.
In Fig. \ref{fig:exp_toy} and Fig. \ref{fig:exp_real}, to predict the future for each agent, SMEMO can focus only on the remaining two agents, since a total of three agents is present in the social context. Therefore, an attention of 0.5 on both agents indicates no relevant interaction.
In such examples, when the attention for an agent over another increases above 0.5, SMEMO has considered such agent relevant for determining the current trajectory and has taken it into account to generate the prediction.

In the example from the SSA dataset (Fig. \ref{fig:exp_toy}), modeling social interactions is necessary to perform adequate predictions.
The agents have to pass through the center of the circle with a precise order, depending on their speed and following the rule described in Sec.~\ref{sec:ssa}: first agent 2, then agent 0 and at last agent 1. When agents get close to each other, i.e. when they need to obey social interaction rules, we can observe spikes in SMEMO's reading attentions.
At timestep 35, agent 0 and 1 both have a very high attention value on agent 2 in order to predict a halt in their prediction. Instead, agent 2 has equal attention on both agents since no counter action is required for the fastest agent, that can proceed along its trajectory, unaffected by the others.
As soon as agent 2 has crossed the center, SMEMO resumes agent 0 while agent 1 still waits. This can be seen at instant 41, when agent 1 has a very high level of attention on agent 0, since it must wait for it to pass before advancing.
A similar analysis can be carried out for the ETH/UCY example (Fig. \ref{fig:exp_real}).
Agents 0 and 1 are in a state of possible collision due to their speed and direction. By observing the attention values at timestep 10 and 12, it can be seen that SMEMO focuses on the second to predict the first and vice-versa, ignoring the behavior of agent 2 which is not relevant in such interaction.
On the other hand, when predicting agent 2's future, there is no relevant peak of attention meaning that its future can be determined without any social information.

Trying to provide a quantitative analysis is challenging. To the best of our knowledge no existing metric to quantitatively evaluate the accuracy of cause-effect relationships in trajectory prediction has been introduced in literature. This stems from the intrinsic difficulty of precisely annotating a dataset with cause-effects, since subjective agent intentions and reactions to the environment should be taken into account.
As an attempt to quantify this phenomena, we leverage the synthetic dataset SSA and the Cause-Effect Accuracy (CEA) defined in Sec. \ref{sec:metrics}. Since motion patterns are synthetically generated, we can explicitly identify cause-effect relationships in social behaviors.
For each timestep where an interaction occurs, i.e. when an agent $i$ is stopped due to another one, we establish the cause of the interaction by taking the agent $j$ with the highest attention value in memory:
$j^* = \arg \max_j \textbf{att}^i(j).$

\minorrevision{Since procedure can in principle be applied to any method that computes an inter-agent attention for each timestep. Thus, we compare the capacity of SMEMO to establish cause-effect relationships against SR-LSTM~\cite{zhang2019sr}, which exploits the attention of a GNN, and AgentFormer~\cite{yuan2021agentformer}, which uses the self-attention of a transformer.
We report the results in Tab. \ref{tab:CEA}.
SMEMO manages to achieve a much higher CEA than the other methods (71\% against 48\% and 39\%).
Interestingly, the three methods are the only ones in Tab.~\ref{tab:toy} performing inter-agent attention for each future timestep and are the top three competitors in terms of ADE, FDE and Kendall. This suggests that modeling relative importance in future timesteps can provide better predictions.
It has to be noted that other methods such as PECNet~\cite{mangalam2020not}, Trajectron++~\cite{salzmann2020trajectron++} and Expert-Goals~\cite{he2021where} perform an inter-agent attention limited to the past or present timesteps and therefore cannot be used for correctly evaluating CEA.}


\begin{table}[t]
	\centering
	\caption{\label{tab:CEA}\minorrevision{Comparison on SSA between methods that use inter-agent attention in every future timesteps: SMEMO, SR-LSTM and AgentFormer. For each method we compare the Cause-Effect Accuracy (CEA).}}
	\begin{tabular}{l|c|c|c|c}
		\multicolumn{1}{c|}{\textbf{Method}} & \multicolumn{1}{l|}{\textbf{ADE}~$\downarrow$} & \multicolumn{1}{l|}{\textbf{FDE}~$\downarrow$} & \textbf{Kendall}~$\uparrow$ & \textbf{CEA}~$\uparrow$\\ \hlineB{3}
  AgentFormer~\cite{yuan2021agentformer} & 0.243 & 0.385 & 0.701 & 0.39 \\
		SR-LSTM~\cite{zhang2019sr}                & 0.217 & 0.409 & 0.777 & 0.48  \\
		SMEMO                              & \textbf{0.169}          & \textbf{0.244}         & \textbf{0.827}  & \textbf{0.71} \\
	\end{tabular}
\end{table}

%
%
%
%
%

%
%
%
%
%

\section{Conclusions}

We present SMEMO, a neural network augmented with a SOcial MEmory MOdule, dealing with the challenging task of multimodal trajectory modeling in social contexts. The algorithmic nature of our approach is able to learn the set of social rules yielding behaviors of pedestrian during their interaction. We created a synthetic dataset to highlight the complex nature of social rule modeling. Finally, we report state-of-the art results for SMEMO on on ETH/UCY and SDD datasets. As a byproduct, we show that SMEMO can provide explainable predictions by design, simply looking at attention weights of its memory reading controllers.

\section*{Acknowledgements}
\scriptsize
This work was supported by the European Commission under European Horizon 2020 Programme, grant number 951911 - AI4Media.

\ifCLASSOPTIONcaptionsoff
  \newpage
\fi



\bibliographystyle{IEEEtran}
\bibliography{egbib}

\begin{thebibliography}{10}
\providecommand{\url}[1]{#1}
\csname url@samestyle\endcsname
\providecommand{\newblock}{\relax}
\providecommand{\bibinfo}[2]{#2}
\providecommand{\BIBentrySTDinterwordspacing}{\spaceskip=0pt\relax}
\providecommand{\BIBentryALTinterwordstretchfactor}{4}
\providecommand{\BIBentryALTinterwordspacing}{\spaceskip=\fontdimen2\font plus
\BIBentryALTinterwordstretchfactor\fontdimen3\font minus \fontdimen4\font\relax}
\providecommand{\BIBforeignlanguage}[2]{{%
\expandafter\ifx\csname l@#1\endcsname\relax
\typeout{** WARNING: IEEEtran.bst: No hyphenation pattern has been}%
\typeout{** loaded for the language `#1'. Using the pattern for}%
\typeout{** the default language instead.}%
\else
\language=\csname l@#1\endcsname
\fi
#2}}
\providecommand{\BIBdecl}{\relax}
\BIBdecl

\bibitem{kothari2020human}
P.~Kothari, S.~Kreiss, and A.~Alahi, ``Human trajectory forecasting in crowds: A deep learning perspective,'' \emph{arXiv preprint arXiv:2007.03639}, 2020.

\bibitem{miyake1999models}
A.~Miyake and P.~Shah, \emph{Models of working memory: Mechanisms of active maintenance and executive control}.\hskip 1em plus 0.5em minus 0.4em\relax Cambridge University Press, 1999.

\bibitem{helbing1995social}
D.~Helbing and P.~Molnar, ``Social force model for pedestrian dynamics,'' \emph{Physical review E}, vol.~51, no.~5, p. 4282, 1995.

\bibitem{pellegrini2009you}
S.~Pellegrini, A.~Ess, K.~Schindler, and L.~Van~Gool, ``You'll never walk alone: Modeling social behavior for multi-target tracking,'' in \emph{2009 IEEE 12th International Conference on Computer Vision}.\hskip 1em plus 0.5em minus 0.4em\relax IEEE, 2009, pp. 261--268.

\bibitem{alahi2016social}
A.~Alahi, K.~Goel, V.~Ramanathan, A.~Robicquet, L.~Fei-Fei, and S.~Savarese, ``Social lstm: Human trajectory prediction in crowded spaces,'' in \emph{Proceedings of the IEEE conference on computer vision and pattern recognition}, 2016, pp. 961--971.

\bibitem{gupta2018social}
A.~Gupta, J.~Johnson, L.~Fei-Fei, S.~Savarese, and A.~Alahi, ``Social gan: Socially acceptable trajectories with generative adversarial networks,'' in \emph{Proceedings of the IEEE Conference on Computer Vision and Pattern Recognition}, 2018, pp. 2255--2264.

\bibitem{ivanovic2019trajectron}
B.~Ivanovic and M.~Pavone, ``The trajectron: Probabilistic multi-agent trajectory modeling with dynamic spatiotemporal graphs,'' in \emph{Proceedings of the IEEE International Conference on Computer Vision}, 2019, pp. 2375--2384.

\bibitem{sadeghian2019sophie}
A.~Sadeghian, V.~Kosaraju, A.~Sadeghian, N.~Hirose, H.~Rezatofighi, and S.~Savarese, ``Sophie: An attentive gan for predicting paths compliant to social and physical constraints,'' in \emph{Proceedings of the IEEE Conference on Computer Vision and Pattern Recognition}, 2019, pp. 1349--1358.

\bibitem{mohamed2020social}
A.~Mohamed, K.~Qian, M.~Elhoseiny, and C.~Claudel, ``Social-stgcnn: A social spatio-temporal graph convolutional neural network for human trajectory prediction,'' in \emph{Proceedings of the IEEE/CVF Conference on Computer Vision and Pattern Recognition}, 2020, pp. 14\,424--14\,432.

\bibitem{graves2014neural}
A.~Graves, G.~Wayne, and I.~Danihelka, ``Neural turing machines,'' \emph{arXiv preprint arXiv:1410.5401}, 2014.

\bibitem{weston2014memory}
J.~Weston, S.~Chopra, and A.~Bordes, ``Memory networks,'' \emph{arXiv preprint arXiv:1410.3916}, 2014.

\bibitem{graves2016hybrid}
A.~Graves, G.~Wayne, M.~Reynolds, T.~Harley, I.~Danihelka, A.~Grabska-Barwi{\'n}ska, S.~G. Colmenarejo, E.~Grefenstette, T.~Ramalho, J.~Agapiou \emph{et~al.}, ``Hybrid computing using a neural network with dynamic external memory,'' \emph{Nature}, vol. 538, no. 7626, pp. 471--476, 2016.

\bibitem{santoro2016meta}
A.~Santoro, S.~Bartunov, M.~Botvinick, D.~Wierstra, and T.~Lillicrap, ``Meta-learning with memory-augmented neural networks,'' in \emph{International conference on machine learning}, 2016, pp. 1842--1850.

\bibitem{marchetti2020memnet}
F.~Marchetti, F.~Becattini, L.~Seidenari, and A.~Del~Bimbo, ``{MANTRA}: Memory augmented networks for multiple trajectory prediction,'' in \emph{Proceedings of the IEEE Conference on Computer Vision and Pattern Recognition}, 2020.

\bibitem{marchetti2020multiple}
F.~Marchetti, F.~Becattini, L.~Seidenari, and A.~D. Bimbo, ``Multiple trajectory prediction of moving agents with memory augmented networks,'' \emph{IEEE Transactions on Pattern Analysis and Machine Intelligence}, vol.~45, no.~6, pp. 6688--6702, 2023.

\bibitem{bach2015pixel}
S.~Bach, A.~Binder, G.~Montavon, F.~Klauschen, K.-R. M{\"u}ller, and W.~Samek, ``On pixel-wise explanations for non-linear classifier decisions by layer-wise relevance propagation,'' \emph{PloS one}, vol.~10, no.~7, p. e0130140, 2015.

\bibitem{selvaraju2017grad}
R.~R. Selvaraju, M.~Cogswell, A.~Das, R.~Vedantam, D.~Parikh, and D.~Batra, ``Grad-cam: Visual explanations from deep networks via gradient-based localization,'' in \emph{Proceedings of the IEEE international conference on computer vision}, 2017, pp. 618--626.

\bibitem{giuliari2020transformer}
F.~Giuliari, I.~Hasan, M.~Cristani, and F.~Galasso, ``Transformer networks for trajectory forecasting,'' \emph{arXiv preprint arXiv:2003.08111}, 2020.

\bibitem{lee2017desire}
N.~Lee, W.~Choi, P.~Vernaza, C.~B. Choy, P.~H. Torr, and M.~Chandraker, ``Desire: Distant future prediction in dynamic scenes with interacting agents,'' in \emph{Proceedings of the IEEE Conference on Computer Vision and Pattern Recognition}, 2017, pp. 336--345.

\bibitem{srikanth2019infer}
S.~Srikanth, J.~A. Ansari, S.~Sharma \emph{et~al.}, ``Infer: Intermediate representations for future prediction,'' in \emph{IEEE/RSJ International Conference on Intelligent Robots and Systems (IROS 2019)}, 2019.

\bibitem{chang2019argoverse}
M.-F. Chang, J.~Lambert, P.~Sangkloy, J.~Singh, S.~Bak, A.~Hartnett, D.~Wang, P.~Carr, S.~Lucey, D.~Ramanan \emph{et~al.}, ``Argoverse: 3d tracking and forecasting with rich maps,'' in \emph{Proceedings of the IEEE Conference on Computer Vision and Pattern Recognition}, 2019, pp. 8748--8757.

\bibitem{caesar2020nuscenes}
H.~Caesar, V.~Bankiti, A.~H. Lang, S.~Vora, V.~E. Liong, Q.~Xu, A.~Krishnan, Y.~Pan, G.~Baldan, and O.~Beijbom, ``nuscenes: A multimodal dataset for autonomous driving,'' in \emph{Proceedings of the IEEE/CVF conference on computer vision and pattern recognition}, 2020, pp. 11\,621--11\,631.

\bibitem{shafiee2021introvert}
N.~Shafiee, T.~Padir, and E.~Elhamifar, ``Introvert: Human trajectory prediction via conditional 3d attention,'' in \emph{Proceedings of the IEEE/CVF Conference on Computer Vision and Pattern Recognition}, 2021, pp. 16\,815--16\,825.

\bibitem{ma2019trafficpredict}
Y.~Ma, X.~Zhu, S.~Zhang, R.~Yang, W.~Wang, and D.~Manocha, ``Trafficpredict: Trajectory prediction for heterogeneous traffic-agents,'' in \emph{Proceedings of the AAAI Conference on Artificial Intelligence}, vol.~33, 2019, pp. 6120--6127.

\bibitem{yuan2021agentformer}
Y.~Yuan, X.~Weng, Y.~Ou, and K.~Kitani, ``Agentformer: Agent-aware transformers for socio-temporal multi-agent forecasting,'' \emph{arXiv preprint arXiv:2103.14023}, 2021.

\bibitem{salzmann2020trajectron++}
T.~Salzmann, B.~Ivanovic, P.~Chakravarty, and M.~Pavone, ``Trajectron++: Multi-agent generative trajectory forecasting with heterogeneous data for control,'' \emph{arXiv preprint arXiv:2001.03093}, 2020.

\bibitem{tang2019multiple}
C.~Tang and R.~R. Salakhutdinov, ``Multiple futures prediction,'' in \emph{Advances in Neural Information Processing Systems}, 2019, pp. 15\,398--15\,408.

\bibitem{mao2023leapfrog}
W.~Mao, C.~Xu, Q.~Zhu, S.~Chen, and Y.~Wang, ``Leapfrog diffusion model for stochastic trajectory prediction,'' in \emph{Proceedings of the IEEE/CVF Conference on Computer Vision and Pattern Recognition}, 2023, pp. 5517--5526.

\bibitem{gu2022stochastic}
T.~Gu, G.~Chen, J.~Li, C.~Lin, Y.~Rao, J.~Zhou, and J.~Lu, ``Stochastic trajectory prediction via motion indeterminacy diffusion,'' in \emph{Proceedings of the IEEE/CVF Conference on Computer Vision and Pattern Recognition}, 2022, pp. 17\,113--17\,122.

\bibitem{deo2018multi}
N.~Deo and M.~M. Trivedi, ``Multi-modal trajectory prediction of surrounding vehicles with maneuver based lstms,'' in \emph{2018 IEEE Intelligent Vehicles Symposium (IV)}.\hskip 1em plus 0.5em minus 0.4em\relax IEEE, 2018, pp. 1179--1184.

\bibitem{lisotto2019social}
M.~Lisotto, P.~Coscia, and L.~Ballan, ``Social and scene-aware trajectory prediction in crowded spaces,'' in \emph{Proceedings of the IEEE International Conference on Computer Vision Workshops}, 2019, pp. 0--0.

\bibitem{Zhao2020TNTTT}
H.~Zhao, J.~Gao, T.~Lan, C.~Sun, B.~Sapp, B.~Varadarajan, Y.~Shen, Y.~Chai, C.~Schmid, C.~Li, and D.~Anguelov, ``Tnt: Target-driven trajectory prediction,'' \emph{ArXiv}, vol. abs/2008.08294, 2020.

\bibitem{Dendorfer_2020_ACCV}
P.~Dendorfer, A.~Osep, and L.~Leal-Taixe, ``Goal-gan: Multimodal trajectory prediction based on goal position estimation,'' in \emph{Proceedings of the Asian Conference on Computer Vision (ACCV)}, November 2020.

\bibitem{mangalam2020not}
K.~Mangalam, H.~Girase, S.~Agarwal, K.-H. Lee, E.~Adeli, J.~Malik, and A.~Gaidon, ``It is not the journey but the destination: Endpoint conditioned trajectory prediction,'' \emph{arXiv preprint arXiv:2004.02025}, 2020.

\bibitem{he2021where}
Z.~He and R.~P. Wildes, ``Where are you heading? dynamic trajectory prediction with expert goal examples,'' in \emph{Proceedings of the International Conference on Computer Vision (ICCV)}, Oct. 2021.

\bibitem{mangalam2021goals}
K.~Mangalam, Y.~An, H.~Girase, and J.~Malik, ``From goals, waypoints \& paths to long term human trajectory forecasting,'' in \emph{Proceedings of the IEEE/CVF International Conference on Computer Vision}, 2021, pp. 15\,233--15\,242.

\bibitem{Kosaraju2019BIGAT}
V.~Kosaraju, A.~Sadeghian, R.~Martin-Martin, I.~Reid, H.~Rezatofighi, and S.~Savarese, ``Social-bigat: Multimodal trajectory forecasting using bicycle-gan and graph attention networks,'' in \emph{Advances in Neural Information Processing Systems}, vol.~32.\hskip 1em plus 0.5em minus 0.4em\relax Curran Associates, Inc., 2019.

\bibitem{shi2021sgcn}
L.~Shi, L.~Wang, C.~Long, S.~Zhou, M.~Zhou, Z.~Niu, and G.~Hua, ``Sgcn: Sparse graph convolution network for pedestrian trajectory prediction,'' in \emph{Proceedings of the IEEE/CVF Conference on Computer Vision and Pattern Recognition}, 2021, pp. 8994--9003.

\bibitem{kothari2021interpretable}
P.~Kothari, B.~Sifringer, and A.~Alahi, ``Interpretable social anchors for human trajectory forecasting in crowds,'' in \emph{Proceedings of the IEEE/CVF Conference on Computer Vision and Pattern Recognition}, 2021, pp. 15\,556--15\,566.

\bibitem{Huang2019stgat}
Y.~Huang, H.~Bi, Z.~Li, T.~Mao, and Z.~Wang, ``Stgat: Modeling spatial-temporal interactions for human trajectory prediction,'' in \emph{2019 IEEE/CVF International Conference on Computer Vision (ICCV)}, 2019, pp. 6271--6280.

\bibitem{hochreiter1997long}
S.~Hochreiter and J.~Schmidhuber, ``Long short-term memory,'' \emph{Neural computation}, vol.~9, no.~8, pp. 1735--1780, 1997.

\bibitem{cho2014learning}
K.~Cho, B.~Van~Merri{\"e}nboer, C.~Gulcehre, D.~Bahdanau, F.~Bougares, H.~Schwenk, and Y.~Bengio, ``Learning phrase representations using rnn encoder-decoder for statistical machine translation,'' \emph{arXiv preprint arXiv:1406.1078}, 2014.

\bibitem{sukhbaatar2015end}
S.~Sukhbaatar, J.~Weston, R.~Fergus \emph{et~al.}, ``End-to-end memory networks,'' in \emph{Advances in neural information processing systems}, 2015, pp. 2440--2448.

\bibitem{rebuffi2017icarl}
S.-A. Rebuffi, A.~Kolesnikov, G.~Sperl, and C.~H. Lampert, ``icarl: Incremental classifier and representation learning,'' in \emph{Proceedings of the IEEE conference on Computer Vision and Pattern Recognition}, 2017, pp. 2001--2010.

\bibitem{yang2018learning}
T.~Yang and A.~B. Chan, ``Learning dynamic memory networks for object tracking,'' in \emph{Proceedings of the European Conference on Computer Vision (ECCV)}, 2018, pp. 152--167.

\bibitem{lai2020mast}
Z.~Lai, E.~Lu, and W.~Xie, ``Mast: A memory-augmented self-supervised tracker,'' in \emph{Proceedings of the IEEE/CVF Conference on Computer Vision and Pattern Recognition}, 2020, pp. 6479--6488.

\bibitem{kumar2016ask}
A.~Kumar, O.~Irsoy, P.~Ondruska, M.~Iyyer, J.~Bradbury, I.~Gulrajani, V.~Zhong, R.~Paulus, and R.~Socher, ``Ask me anything: Dynamic memory networks for natural language processing,'' in \emph{International conference on machine learning}, 2016, pp. 1378--1387.

\bibitem{ma2018visual}
C.~Ma, C.~Shen, A.~Dick, Q.~Wu, P.~Wang, A.~van~den Hengel, and I.~Reid, ``Visual question answering with memory-augmented networks,'' in \emph{Proceedings of the IEEE Conference on Computer Vision and Pattern Recognition}, 2018, pp. 6975--6984.

\bibitem{pernici2020self}
F.~Pernici, M.~Bruni, and A.~Del~Bimbo, ``Self-supervised on-line cumulative learning from video streams,'' \emph{Computer Vision and Image Understanding}, p. 102983, 2020.

\bibitem{han2020memory}
T.~Han, W.~Xie, and A.~Zisserman, ``Memory-augmented dense predictive coding for video representation learning,'' \emph{arXiv preprint arXiv:2008.01065}, 2020.

\bibitem{DBLP:conf/icpr/DivitiisBBB20}
\BIBentryALTinterwordspacing
L.~D. Divitiis, F.~Becattini, C.~Baecchi, and A.~D. Bimbo, ``Garment recommendation with memory augmented neural networks,'' in \emph{Pattern Recognition. {ICPR} International Workshops and Challenges - Virtual Event, January 10-15, 2021, Proceedings, Part {II}}, ser. Lecture Notes in Computer Science, vol. 12662.\hskip 1em plus 0.5em minus 0.4em\relax Springer, 2020, pp. 282--295. [Online]. Available: \url{https://doi.org/10.1007/978-3-030-68790-8\_23}
\BIBentrySTDinterwordspacing

\bibitem{marchetti2022explainable}
F.~Marchetti, F.~Becattini, L.~Seidenari, and A.~Del~Bimbo, ``Explainable sparse attention for memory-based trajectory predictors,'' in \emph{European Conference on Computer Vision}.\hskip 1em plus 0.5em minus 0.4em\relax Springer, 2022, pp. 543--560.

\bibitem{xu2022remember}
C.~Xu, W.~Mao, W.~Zhang, and S.~Chen, ``Remember intentions: Retrospective-memory-based trajectory prediction,'' in \emph{Proceedings of the IEEE/CVF Conference on Computer Vision and Pattern Recognition}, 2022, pp. 6488--6497.

\bibitem{yang2022continual}
B.~Yang, F.~Fan, R.~Ni, J.~Li, L.~Kiong, and X.~Liu, ``Continual learning-based trajectory prediction with memory augmented networks,'' \emph{Knowledge-Based Systems}, vol. 258, p. 110022, 2022.

\bibitem{fernando2018tree}
T.~Fernando, S.~Denman, A.~McFadyen, S.~Sridharan, and C.~Fookes, ``Tree memory networks for modelling long-term temporal dependencies,'' \emph{Neurocomputing}, vol. 304, pp. 64--81, 2018.

\bibitem{yu2020spatio}
C.~Yu, X.~Ma, J.~Ren, H.~Zhao, and S.~Yi, ``Spatio-temporal graph transformer networks for pedestrian trajectory prediction,'' in \emph{European Conference on Computer Vision}.\hskip 1em plus 0.5em minus 0.4em\relax Springer, 2020, pp. 507--523.

\bibitem{li2022graph}
L.~Li, M.~Pagnucco, and Y.~Song, ``Graph-based spatial transformer with memory replay for multi-future pedestrian trajectory prediction,'' in \emph{Proceedings of the IEEE/CVF Conference on Computer Vision and Pattern Recognition}, 2022, pp. 2231--2241.

\bibitem{deo2018convolutional}
N.~Deo and M.~M. Trivedi, ``Convolutional social pooling for vehicle trajectory prediction,'' in \emph{Proceedings of the IEEE Conference on Computer Vision and Pattern Recognition Workshops}, 2018, pp. 1468--1476.

\bibitem{berlincioni2020multiple}
L.~Berlincioni, F.~Becattini, L.~Seidenari, and A.~Del~Bimbo, ``Multiple future prediction leveraging synthetic trajectories,'' 2020.

\bibitem{buhet2019conditional}
T.~Buhet, E.~Wirbel, and X.~Perrotton, ``Conditional vehicle trajectories prediction in carla urban environment,'' in \emph{Proceedings of the IEEE/CVF International Conference on Computer Vision Workshops}, 2019, pp. 0--0.

\bibitem{Robicquet2016sdd}
A.~Robicquet, A.~Sadeghian, A.~Alahi, and S.~Savarese, ``Learning social etiquette: Human trajectory understanding in crowded scenes,'' in \emph{Computer Vision -- ECCV 2016}, B.~Leibe, J.~Matas, N.~Sebe, and M.~Welling, Eds.\hskip 1em plus 0.5em minus 0.4em\relax Cham: Springer International Publishing, 2016, pp. 549--565.

\bibitem{sadeghian2018trajnet}
A.~Sadeghian, V.~Kosaraju, A.~Gupta, S.~Savarese, and A.~Alahi, ``Trajnet: Towards a benchmark for human trajectory prediction,'' \emph{arXiv preprint}, 2018.

\bibitem{pellegrini2010eth}
S.~Pellegrini, A.~Ess, and L.~Van~Gool, ``Improving data association by joint modeling of pedestrian trajectories and groupings,'' in \emph{Computer Vision -- ECCV 2010}, K.~Daniilidis, P.~Maragos, and N.~Paragios, Eds.\hskip 1em plus 0.5em minus 0.4em\relax Berlin, Heidelberg: Springer Berlin Heidelberg, 2010, pp. 452--465.

\bibitem{lerner2007ucy}
A.~Lerner, Y.~Chrysanthou, and D.~Lischinski, ``Crowds by example,'' \emph{Comput. Graph. Forum}, vol.~26, pp. 655--664, 09 2007.

\bibitem{kendall1948rank}
M.~G. Kendall, ``Rank correlation methods.'' 1948.

\bibitem{zhang2019sr}
P.~Zhang, W.~Ouyang, P.~Zhang, J.~Xue, and N.~Zheng, ``Sr-lstm: State refinement for lstm towards pedestrian trajectory prediction,'' in \emph{Proceedings of the IEEE/CVF Conference on Computer Vision and Pattern Recognition}, 2019, pp. 12\,085--12\,094.

\bibitem{li2020evolvegraph}
J.~Li, F.~Yang, M.~Tomizuka, and C.~Choi, ``Evolvegraph: Multi-agent trajectory prediction with dynamic relational reasoning,'' \emph{Proceedings of the Neural Information Processing Systems (NeurIPS)}, 2020.

\bibitem{ridel2020scene}
D.~{Ridel}, N.~{Deo}, D.~{Wolf}, and M.~{Trivedi}, ``Scene compliant trajectory forecast with agent-centric spatio-temporal grids,'' \emph{IEEE Robotics and Automation Letters}, vol.~5, no.~2, pp. 2816--2823, 2020.

\bibitem{sun2021three}
J.~Sun, Y.~Li, H.-S. Fang, and C.~Lu, ``Three steps to multimodal trajectory prediction: Modality clustering, classification and synthesis,'' \emph{arXiv preprint arXiv:2103.07854}, 2021.

\bibitem{pang2021trajectory}
B.~Pang, T.~Zhao, X.~Xie, and Y.~N. Wu, ``Trajectory prediction with latent belief energy-based model,'' in \emph{Proceedings of the IEEE/CVF Conference on Computer Vision and Pattern Recognition}, 2021, pp. 11\,814--11\,824.

\bibitem{bhattacharyya2020conditional}
A.~Bhattacharyya, M.~Hanselmann, M.~Fritz, B.~Schiele, and C.-N. Straehle, ``Conditional flow variational autoencoders for structured sequence prediction,'' 2020.

\bibitem{deo2020trajectory}
N.~Deo and M.~M. Trivedi, ``Trajectory forecasts in unknown environments conditioned on grid-based plans,'' \emph{arXiv preprint arXiv:2001.00735}, 2020.

\bibitem{liang2020simaug}
J.~Liang, L.~Jiang, and A.~Hauptmann, ``Simaug: Learning robust representations from simulation for trajectory prediction,'' in \emph{European Conference on Computer Vision}.\hskip 1em plus 0.5em minus 0.4em\relax Springer, 2020, pp. 275--292.

\bibitem{liang2019peeking}
J.~Liang, L.~Jiang, J.~C. Niebles, A.~G. Hauptmann, and L.~Fei-Fei, ``Peeking into the future: Predicting future person activities and locations in videos,'' in \emph{Proceedings of the IEEE/CVF Conference on Computer Vision and Pattern Recognition}, 2019, pp. 5725--5734.

\bibitem{li2019cgns}
J.~Li, H.~Ma, and M.~Tomizuka, ``Conditional generative neural system for probabilistic trajectory prediction,'' 11 2019, pp. 6150--6156.

\bibitem{zhao2019multi}
T.~Zhao, Y.~Xu, M.~Monfort, W.~Choi, C.~Baker, Y.~Zhao, Y.~Wang, and Y.~N. Wu, ``Multi-agent tensor fusion for contextual trajectory prediction,'' in \emph{Proceedings of the IEEE/CVF Conference on Computer Vision and Pattern Recognition}, 2019, pp. 12\,126--12\,134.

\bibitem{liu2021social}
Y.~Liu, Q.~Yan, and A.~Alahi, ``Social nce: Contrastive learning of socially-aware motion representations,'' in \emph{Proceedings of the IEEE/CVF International Conference on Computer Vision}, 2021, pp. 15\,118--15\,129.

\end{thebibliography}
%
%
%

%

\begin{IEEEbiography}[{\includegraphics[width=1in,height=1.25in,clip,keepaspectratio]{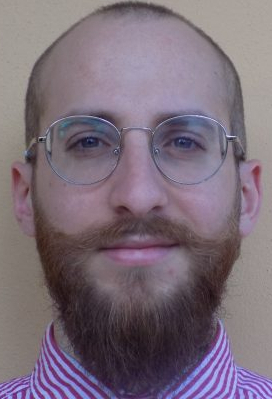}}]{Francesco Marchetti} received a master degree  cum laude in 2019 in computer engineering from the University of Florence with the thesis “Trajectories Prediction for Autonomous Driving with Memory Networks” in collaboration with the research institute IMRA Europe.
Currently he is a PhD student at Media Integration and Communication Center (MICC) and the research work focuses on trajectories forecasting in the automotive field.
\end{IEEEbiography}

\begin{IEEEbiography}[{\includegraphics[width=1in,height=1.25in,clip,keepaspectratio]{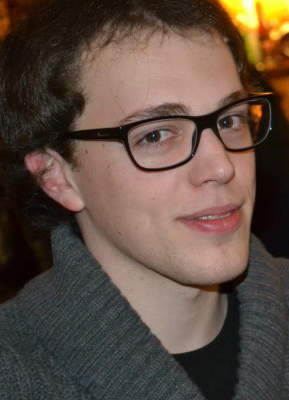}}]	{Federico Becattini} obtained his PhD in 2018 from the University of Florence under the supervision of Prof. Alberto Del Bimbo and Prof. Lorenzo Seidenari. Currently he is a Tenure Track Assistant Professor at the University of Siena. His research interest are Autonomous Driving and Scene Understanding. He served to the scientific community as a reviewer for scientific journals and conferences and has organized workshops and tutorials at international venues.	
\end{IEEEbiography}

\begin{IEEEbiography}
[{\includegraphics[width=1in,height=1.25in,clip,keepaspectratio]{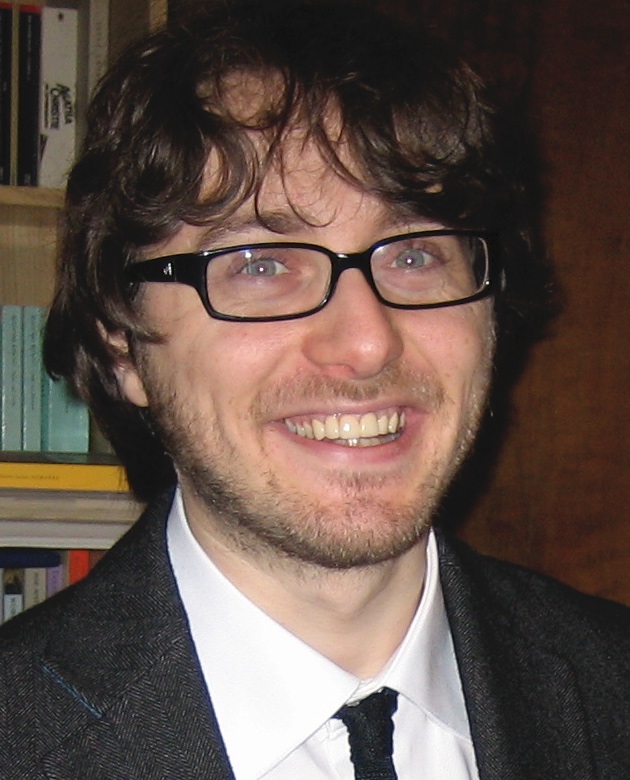}}]{Lorenzo Seidenari}
is an Assistant Professor at the Department of Information Engineering of the  University of Florence. He received his Ph.D. degree in computer engineering in 2012 from the University of Florence. His research focuses on deep learning for object and action recognition in video and images.  He is an ELLIS scholar.   He is author of 16 journal papers and more than 40 peer-reviewed conference papers. He has an h-index of 25 with more than 2200 citations. 
\end{IEEEbiography}

\begin{IEEEbiography}[{\includegraphics[width=1in,height=1.25in,clip,keepaspectratio]{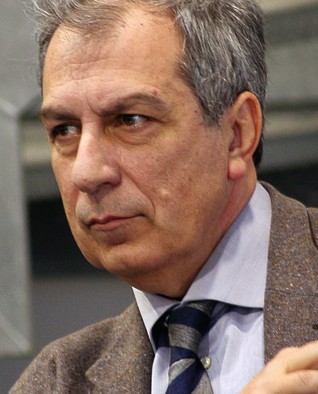}}]{Alberto Del Bimbo}is a Full Professor of Computer Engineering. His scientific interests are multimedia information retrieval, pattern recognition, image and video analysis, and human–computer interaction. From 1996 to 2000, he was the President of the IAPR Italian Chapter and the Member-at-Large of the IEEE Publication Board from 1998 to 2000. He was the General Co-Chair of ACMMM2010 and ECCV2012. He was nominated as ACM Distinguished Scientist in 2016. He received the SIGMM Technical Achievement Award for Outstanding Technical Contributions to Multimedia Computing, Communications and Applications. He is an IAPR Fellow, and an Associate Editor of several international journals.
\end{IEEEbiography}




\end{document}